\documentclass{article}
\usepackage{setspace,caption}

\usepackage[utf8]{inputenc}
\usepackage{hyperref}

\usepackage{amssymb}
\usepackage{mathtools}
\usepackage{graphicx}
\usepackage{booktabs}
\usepackage{stmaryrd}
\usepackage{color}
\usepackage{todonotes}
\usepackage{multirow}
\usepackage{amsmath,amsfonts,bm}

\newcommand{\newterm}[1]{{\bf #1}}

\def\eqref#1{equation~\ref{#1}}
\def\1{\bm{1}}

\def\vd{{\bm{d}}}
\def\ve{{\bm{e}}}

\def\vh{{\bm{h}}}

\def\vl{{\bm{l}}}

\def\vr{{\bm{r}}}
\def\vs{{\bm{s}}}

\def\vu{{\bm{u}}}
\def\vv{{\bm{v}}}

\def\vx{{\bm{x}}}
\def\vy{{\bm{y}}}

\def\evlambda{{\lambda}}

\def\mL{{\bm{L}}}
\def\mM{{\bm{M}}}

\def\mP{{\bm{P}}}

\def\mR{{\bm{R}}}

\def\mW{{\bm{W}}}
\def\mX{{\bm{X}}}

\def\mLambda{{\bm{\Lambda}}}
\def\mSigma{{\bm{\Sigma}}}

\DeclareMathAlphabet{\mathsfit}{\encodingdefault}{\sfdefault}{m}{sl}
\SetMathAlphabet{\mathsfit}{bold}{\encodingdefault}{\sfdefault}{bx}{n}

\def\emW{{W}}

\newcommand{\R}{\mathbb{R}}

\def\vlambda{{\bm{\lambda}}}
\def\vnu{{\bm{\nu}}}
\renewcommand{\a}{\alpha}
\renewcommand{\b}{\beta}

\newcommand{\w}{\omega}
\newcommand{\mw}{\omega}

\renewcommand{\H}{\bm{H}}
\newcommand{\vdelta}{\bm{\delta}}
\renewcommand{\R}[1]{\lfloor #1 \rfloor_+}

\newcommand{\trs}{^{\dagger}}
\graphicspath{{figs/}}

\usepackage[authoryear]{natbib}

\usepackage{rotating}
\usepackage{bbm}
\usepackage{latexsym}

\newcommand{\captionfonts}{\normalsize}

\makeatletter
\long\def\@makecaption#1#2{%
  \vskip\abovecaptionskip
  \sbox\@tempboxa{{\captionfonts #1: #2}}%
  \ifdim \wd\@tempboxa >\hsize
    {\captionfonts #1: #2\par}
  \else
    \hbox to\hsize{\hfil\box\@tempboxa\hfil}%
  \fi
  \vskip\belowcaptionskip}
\makeatother
\begin{document}

% \linenumbers

\title{Low-Dimensional Manifolds Support Multiplexed\\ Integrations in~Recurrent~Neural~Networks}
\author{Arnaud Fanthomme \&  R\'emi Monasson \\
   Laboratoire de Physique de l'Ecole Normale Sup\'erieure\\
   PSL \& CNRS UMR 8023, Sorbonne Universit\'e\\
   24 rue Lhomond, 75005 Paris, Paris, France\\

}
\date{\today}

\maketitle

\begin{abstract}
\normalsize

We study the learning dynamics and the representations emerging in Recurrent Neural Networks trained to integrate one or multiple temporal signals.
%We introduce a general loss function, adaptable to any non linearity in the neural activation, that can be used to train networks without evaluating the network on any input data.
Combining analytical and numerical investigations, we characterize the conditions under which a RNN with $n$ neurons learns to integrate $D (\ll n)$ scalar signals of arbitrary duration. We show, for linear, ReLU and sigmoidal neurons, that the internal state lives close to a $D$-dimensional manifold, whose shape is related to the activation function.
Each neuron therefore carries, to various degrees, information about the value of all integrals. We discuss the deep analogy between our results and the concept of \textit{mixed selectivity} forged by computational neuroscientists to interpret cortical recordings.


\end{abstract}

%%%%
%
% Intro
%
%%%%

\section{Introduction}
\label{sec:intro}

Recurrent neural networks (RNNs) have emerged over the past years as a versatile and powerful architecture for supervised learning of complex tasks from examples, involving in particular dynamical processing of temporal signals \citep{EmpiricalRNN}. Applications of RNNs or of their variants designed to capture very long-term dependencies in input sequences through gating mechanisms, such as GRU or LSTM, are numerous and range from state-of-the-art speech recognition networks \citep{DeepSpeech} to protein sequence analysis \citep{DeepLoc}.

How these tasks are actually learned and performed has been extensively studied in the Reservoir Computing setup where the recurrent part of the dynamics is fixed, see \citep{ReservoirReview} for a review, while the general case of RNNs remains mostly an open question. Understanding of those networks would bring valuable advantages to both neuroscience and machine learning, as suggested in \citep{barak_recurrent_2017, richards_deep_2019}. Some results have been recently obtained, when the representations and the dynamics are low dimensional  \citep{sussillo_opening_2012, LowRankRNNs, schuessler_dynamics_2020, schuessler_interplay_2020}, a prominent feature of the Neural Integrators that are the focus of the present study. Neural integrators, whose function is to perform integration of time-series, have been studied for several decades in neuroscience, both experimentally \citep{NeuralIntegratorRobinson, GoldfishVPNI, WangDecisionMaking} and theoretically \citep{Elman_RNN, SeungLinear, SeungNonLinear}, and more recently, numerically, in the context of machine learning \citep{SongWangEINets}.

The goal of our study is three-fold. First, we want to study how exactly the task of integration is learned from examples by RNNs. We derive rigorous results for linear RNNs and approximate ones for non-linear RNNs,  which can be compared to numerical simulations. This approach is similar to the one adopted by \citep{saxe_exact_2014} for the case of deep networks, and more recently by \citep{schuessler_interplay_2020} in the case of recurrent networks. Second, we seek to understand the nature of the internal representations built during learning, an issue of general interest in neural network-based learning \citep{li_understanding_2017, InterpretingZhang, InterpretingMontavon, InterpretingInteractive}. It is, in particular, tempting to compare representations emerging in artificial neural networks to  their natural counterparts in computational neuroscience.  Third, we do not limit ourselves to a single integration, but consider the issue of learning multiple integrators within a unique network. While we do not expect an increase in performance for each individual task, as was observed in the case of Natural Language Processing by \citep{MultitaskSequenceSequence}, we are interested in finding representations adequate for parallel computations within a single network, allowing for considerations on the topic of "mixed selectivity" developed in computational neuroscience and studied by \citep{MixedSelectivity}. The issue of network capacity, the maximum number of tasks that can be performed in parallel, has been previously studied numerically by \citep{collins_capacity_2017}, but remains out of the scope of this study, which will focus on a number $D$ of integrals small compared to the network size.

Our paper is organized as follows. We define the RNNs we consider in this work, the integration task and the training procedure in Section \ref{sec:model}. The case of linear activation function is studied in detail in Section \ref{sec:linear}. RNNs with non-linear activation functions are studied in Section \ref{sec:single_channel} in the case of a single channel ($D=1$), while our results for the general situation of multiple channels ($D\ge 2$) are presented in Section \ref{sec:multi_channel}. Conclusions and perspectives can be found in Section \ref{sec:conclusion}. The paper is complemented by a series of Appendices containing details about the calculations, simulations and further figures. The source code for the simulations can be found at \url{https://github.com/AFanthomme/ManifoldsSupportRNI}.

\begin{figure}[t]
   \centering{\includegraphics[width=\textwidth]{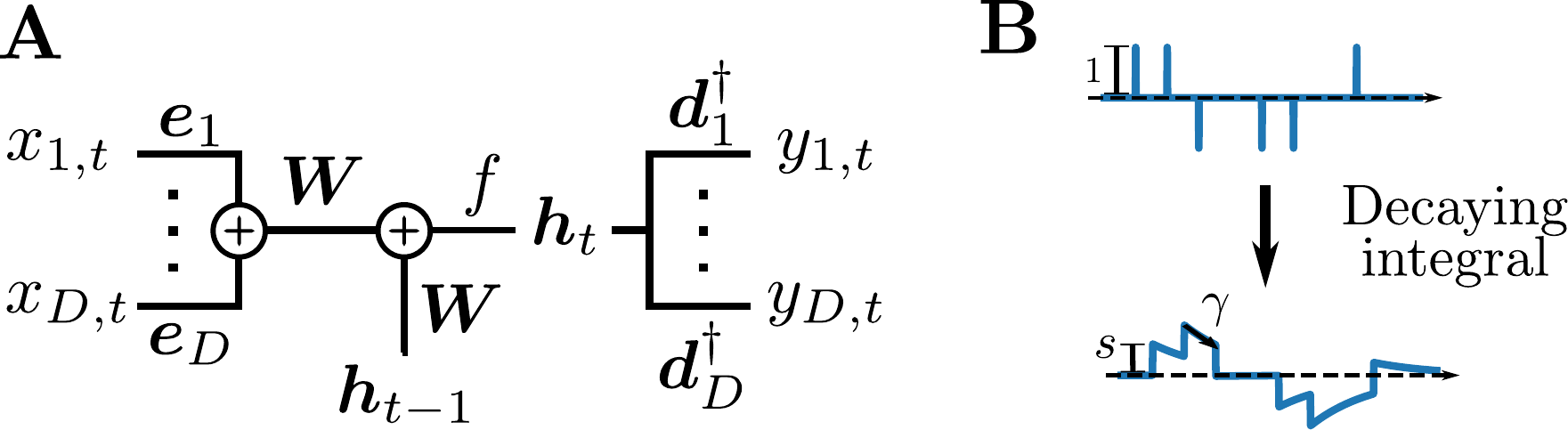}
    \caption{\textbf{A}: Multiplexed Recurrent Neural Network, with $D$ input channels (left) and the same number of output channels (right). The internal state of the RNN, $\vh_t$, is a vector of dimension $n$. The inputs are encoded by the vectors $\ve_c$ and decoded from the internal state through the decoder weights $\vd_c$. \textbf{B}: Illustration of the decaying integral mapping that we want networks to approximate, on a sparse input sequence. At each time-step $t$, if the input time-series $x_t$ is non-zero, the integral is increased by $s \, x_t$; then, it is multiplied by $\gamma<1$, which produces the exponential decay in absence of inputs. In practice, sequences used for experiments were Gaussian noise, and these sparse sequences are used only for visualization.}
    \label{fig:network_and_task}}
\end{figure}

%%%%
%
% Definitions and Model
%
%%%%

\section{Definitions and model}
\label{sec:model}

\paragraph{Description of the network. }
We consider a single-layer %, multiplexed
RNN of size $n$, without any gating mechanism; while such refinements are found to improve performance, as reviewed by \citep{RNNReview}, we omit them as they are not necessary for such a simple task. The computation diagram presented in Figure~\ref{fig:network_and_task}A can be summed up as follows: at time $t$, the scalar inputs along all channels $c=1..D$, denoted $x_{c,t}$,
are multiplied by their respective \newterm{encoder} vectors $\ve_c$; these vectors are summed to the previous internal state $\vh_{t-1}$\footnote{  We initialize the internal state before any input to $\vh_{-1}=\bm{0}$.}, and multiplied by the \newterm{weight matrix} $\mW$ before a componentwise  \newterm{activation} $f$ is applied to get the new internal state $\vh_{t}$. The  update equation for $\vh$ is therefore
\begin{equation}
    \vh_{t} = f\left(\vnu_{t}\right) \ ,
\end{equation}
where the current\footnote{ This name is chosen in analogy with computational neuroscience, where $W_{ij}$ is a synaptic weight from neuron $j$ to neuron $i$, and the image of the activity vector, $\nu_i=\sum_j W_{ij} h_j$, represents the total current from the recurrent population coming onto neuron $i$.} $\vnu_t$ is defined through
\begin{equation}\label{defcurr}
    \vnu_{t} = \mW \cdot \left[\vh_{t-1} + \sum_{c=1}^D x_{c, t}\, \ve_c\right]\ .
\end{equation}
The output units are linear: their values $y_{c,t}$ are simply obtained by taking the scalar product of $\vh_{t}$ and the \newterm{decoder} vectors $\vd_c$,
\begin{equation}
y_{c,t} =\vd_c\cdot  \vh_{t}\ .
\end{equation}

% In the following we will consider two different activation functions $f$: the "linear" activation which is simply the identity, and the ReLU non-linearity, which takes componentwise maximum of a vector and 0.

Most of this study will be focused on two different activation functions $f$: the "linear" activation, which is simply the identity, and the ReLU non-linearity, which takes component-wise maximum of a vector and zero. Linear activation allows for exact results to be derived on both the learning dynamics and the structure of solutions, while the choice of ReLU will serve as an example of non-linear activation that can be used to create perfectly generalizing integrators (at least in the $D=1$ case), and show that the conclusions of the linear network study remain relevant. Finally, we propose a generic procedure to train a RNN with arbitrary non-linearity $f$ to perform multiplexed integrations, which we illustrate with success in the case of sigmoidal activation.

\paragraph{Description of the task.}
The networks will be trained to map $D$ input time-series $(x_{c, t})_{t\in \mathbb{N}}$ to $D$ output ones $(y_{c, t})_{t\in \mathbb{N}}$: for all channels $c=1,...,D$, the $c$-th output should match the $\gamma_c$-discounted sum of the $c$-th channel inputs, times the scale factor $s_c$:
\begin{equation} \label{target1}
\overline{y}_{c,t} =s_c \sum_{k=0}^t \gamma_c^{k+1}   \,  x_{c, t-k} \ ,
\end{equation}
see Figure~\ref{fig:network_and_task}B. The values of the decay constants $\gamma_c$ are chosen in $[0,1]$ to restrict memory to recent events and avoid instabilities\footnote{ If $\gamma$ is chosen too close to $1$, the network might during training have an effective "decay" larger than 1; in that case, the values of the outputs and the associated gradients become large (in particular when training on long input sequences), which can then be overcompensated and make the training divergent.}.

We quantify the performance of the network through the mean square error between the actual and target outputs across the $D$ channels on training epochs of length (duration) $T$:
\begin{equation}\label{loss1}
\mathcal{L}=\left\langle \sum_{c=1}^{D}\sum_{t=0}^{T-1}\big(y_{c,t}-\overline{y}_{c,t}\big)^2  \right \rangle_X .
\end{equation}

\paragraph{Description of the learning procedure.}
% As we are interested in the network's dynamical evolution during the learning phase,
Except when otherwise specified, the encoder $\bf e$ and decoder $\bf d$ will be considered as randomly fixed at network initialization, and forced to be of unit norm. The reason for this hypothesis is two-fold. First, our focus of interest is how the network of connections between neurons evolves during training and the nature of the solutions and representations obtained. The simplified setup allows for deeper mathematical analysis of the dynamics of the $\bf W$ than the general case, where all parameters of the network evolve simultaneously during training. Second, while the speed of convergence is positively impacted by relaxing the constraint of fixing the decoder, numerical experiments indicate that the nature of the $\bf W$ network is qualitatively unchanged if $\bf e$ and $\bf d$ are also trained, in particular when it comes to the way the integrals are represented.

For theoretical analysis, we train the recurrent weights $\mW$ using Gradient Descent (GD) updates at learning rate $\eta$:
\begin{equation}\label{dyna9}
    W_{ij}^{(\tau+1)} = W_{ij}^{(\tau)} - \eta\, \frac{\partial \mathcal{L}}{\partial W_{ij}}(\mW^{(\tau)}) \ ,
\end{equation}
where $\tau$ is the discrete learning time. We also performed experiments using the non-linear Adam optimizer \citep{Adam} to ensure robustness of our results with respect to the specific choice of optimization procedure. Numerical implementations were performed in Python, making extensive use of the Scipy \citep{scipy} and Pytorch \citep{pytorch} packages respectively for scientific computing and implementation of Automatic Differentiation and Gradient Descent optimization.

%%%%
%
% Linear
%
%%%%

\section{Case of linear activation}
\label{sec:linear}

Throughout this section we assume that the activation function $f$ is linear. We start with the simplest case of a single channel ($D=1$), and will omit the subscript $c=1$ below for simplicity; the case of multiple channels $D\ge 2$ will be studied in Section \ref{sec:linear2}.

As the network dynamics $\vh_{t}\to \vh_{t+1}$ is linear, the loss (\ref{loss1}) can be analytically averaged over the input data distribution. The computation is presented in Appendix \ref{app:linear_loss}, and yields:
\begin{equation}
\label{eq:linear_avg_loss}
\mathcal{L}(\mW) = \sum_{q, p=1}^{T} \bm{\chi}_{qp}(\mu_{q}-s\gamma^{q})(\mu_{p}-s\gamma^{p}),
\end{equation}
where
\begin{equation}
\mu_q = \vd\trs \mW^{q}\ve
\end{equation}
 will be hereafter referred to as the $q$-th moment of $\mW$, and $\bm{\chi}$ is a positive-definite matrix, related to the covariance matrix of the inputs $x_t$.

The average loss implicitly depends on $\gamma$, $T$, $s$, $\ve$, $\vd$ and input correlations $\bm{\chi}$, which do not evolve during training and are therefore omitted from the argument. Since $\bm{\chi}$ is positive-definite, the global minimum of the loss is reached when the moments of $\mW$ fulfill
\begin{equation}
\label{eq:linear_avg_minimization_conditions}
    \mu_{q} = s \gamma^{q} \ ,
\end{equation}
for all $q =1, \ldots, T$. The same conditions are obtained for uncorrelated inputs, so we will restrict to this case for numerical investigations in the following.

The gradient of the averaged loss with respect to the weight matrix $\mW$ can be computed (see Appendix \ref{app:gradient_hessian_linear}), with the result
\begin{equation}
    \frac{\partial \mathcal{L}}{\partial \emW_{ij}} = 2  {\sum_{q,p=1}^{T}} \bm{\chi}_{qp}\, (\mu_q\!-\!s\gamma^{q}) \, {\sum_{m=0}^{p-1}}\sum_{\a=1}^{n} d_{\alpha}(W^m)_{\alpha i} \sum_{\b=1}^{n} (W^{p-1-m})_{j \beta}\, e_{\beta}\ .
\label{eq:linear_gradient}
\end{equation}
We emphasize that, while the network update dynamics is linear, the training dynamics over $\mW$ defined by (\ref{dyna9}) and (\ref{eq:linear_gradient}) is highly non-linear.

\subsection{Conditions for generalizing integrators}
\label{sec:generalizing_minima}
Conditions  (\ref{eq:linear_avg_minimization_conditions}) over the moments $\mu_q$, with $q=1,...,T$, ensure that the RNN will perfectly integrate input sequences of length up to the epoch duration $T$. We call \newterm{generalizing integrator (GI)} a RNN such that these conditions are satisfied for {\bf all} integer-valued $q$, ensuring perfect integration of input sequences of arbitrary length.

We will now derive a set of sufficient and necessary conditions for a diagonalizable matrix $\mW$ to be a GI\footnote{
%By the Jordan-Chevalley decomposition, one can always write $\mW$ as a sum of two matrices, one diagonalizable and one nilpotent, such that, for $T$ large enough, the behaviour of the full matrix and of the diagonalizable part will coincide. Additionally,
As the set of diagonalizable matrices is dense in the space of matrices, any non-diagonalizable matrix $\mW$ can be made diagonalizable through the addition of an infinitesimal matrix; the moments of the resulting matrix are arbitrarily close to the ones of $\mW$, which makes our results for diagonalizable matrices directly applicable to $\mW$, see Section \ref{sec:linear_full_rank_link}.}. Let us assume $\mW$ is diagonalized as $\mP \mLambda \mP^{-1}$, where the spectral matrix $\mLambda=\text{diag}(\vlambda)$ is diagonal and $\mP$ is invertible, of inverse $\mP^{-1}$. The moments of $\mW$ can be expressed from the eigenvalues as follows:
\begin{equation*}
    \mu_q = \vd\trs \mP \mLambda^q \mP^{-1}\ve = \sum_{i=1}^n g_i \evlambda_i^q \quad \text{with} \quad g_i= (\mP\trs \vd)_i (\mP^{-1} \ve)_i \ .
\end{equation*}

Obviously, a null eigenvalue does not contribute to the above sum, hence the conditions that we obtain in the following will only apply to non-zero eigenvalues.
Our condition for null loss is that all of the aforementioned moments $\mu_q$ are equal to $s\, \gamma^q$.

The above set of conditions can be rewritten as follows. For any real-valued polynomial $Q(z)$ of degree less than, or equal to $T$ in $z$, such that $Q(0)=0$, we have
\begin{equation}\label{egalQ}
\sum_i g_i \, Q(\lambda_i) = s\, Q(\gamma)  \ .
\end{equation}

We can evaluate the previous equality for well-chosen polynomials. Let us consider one eigenvalue, say, $\evlambda_{\kappa}$ assumed to be different from $\gamma$, and the Lagrange Polynomial $Q(z)$ equal to one for $z=\evlambda_{\kappa}$ and to $0$ for $z=0$, $z=\evlambda_i \ne \lambda_\kappa$ and $z=\gamma$. Such a polynomial exists as soon as $T\geq n+1$ in the general case where all eigenvalues are distinct from each other, $0$, $\gamma$, and as soon as $T\geq r+1$ if $n-r$ eigenvalues are equal \textit{e.g.} to 0. Equality (\ref{egalQ}) gives:
\begin{equation*}
    \sum_i g_i\, \delta_{\evlambda_i, \evlambda_{\kappa}} = 0\ ,
\end{equation*}
where $\delta_{\cdot,\cdot}$ denotes the Kronecker delta.
Therefore, any eigenvalue  different from $\gamma$ must satisfy an exact cancellation condition for the associated $g$ coefficients ensuring that it does not contribute to the network output. Similarly, a condition for the $\gamma$ eigenvalue can be written, to ensure that an input of magnitude $1$ entails a change of magnitude $s$ in the output.

The necessary and sufficient conditions for a diagonalizable matrix $\mW$ to be a global minimum of the loss defined with $T\geq n+1$ therefore read
\begin{equation}
\label{eq:null_loss_conds}
\begin{cases}
&\sum_i g_i\,\delta_{\lambda_i, \gamma} = s \\
\forall \kappa \text{ s.t. } \lambda_{\kappa} \notin \{\gamma, 0\}, \hspace{0.2cm} &\sum_i g_i \,\delta_{\lambda_i, \lambda_{\kappa}} = 0
\end{cases}
\end{equation}

These conditions are in turn enough to guarantee that $\mW$ is a global minimum of the loss for any value of $T$, hence the Generalizing Integrators and the minima of the losses defined with $T\geq n+1$ are equal.

Clearly, any global minimum of the averaged loss $\mathcal{L}$ experimentally obtained when using training sequenes of length $T\geq n+1$ is a GI. Networks trained with much shorter epochs can also be GIs if the rank of $\mW$ remains small enough throughout the training dynamics. More precisely, if we assume we have found a minimum of the loss of rank $r\leq n$ it will be a GI as soon as $T\geq r+1$. An important illustration is provided by the null initialization of the weights ($\mW^{(\tau=0)}=\bm{0}$), which ensures that $\mW$ remains of rank $r=2$ at all times $\tau$, see (\ref{eq:linear_gradient}) and next subsection.

\subsection{Special case of null-weight initialization}
\label{sec:linear_null_init}

We now assume that the weight matrix $\mW$ is initially set to zero, and
characterize all the GIs accessible through GD, as well as the local convergence to those solutions. A  study of the full training dynamics for two special cases ($T=1$ and $\ve=\vd$) can be found in Appendix \ref{app:special_cases}.

\paragraph{Low-rank parametrization.}
From the expression of the gradients (\ref{eq:linear_gradient}) and the linearity of the weight updates (\ref{dyna9}), it is clear that  starting from $\mW\!=\!\bf{0}$, the weight matrix will remain at all times $\tau$ in the subspace generated by the four rank-1 matrices $\vd \vd\trs, \vd \ve\trs, \ve \vd\trs, \ve \ve\trs$. We introduce an orthonormal basis for the $\vv_1\equiv \ve,\vv_2\equiv \vd$ space,
\begin{equation}
\overline{\vv_a} = \sum_{b=1}^{2} (\mSigma^{-1/2})_{ab} \, \vv_b, \quad \text{with} \quad \mSigma_{ab}=\vv_a\trs \vv_b \ ,
\end{equation}
and the corresponding parametrization of the subspace spanned by $\mW$:
\begin{equation}\label{omegatoW}
\mW^{(\tau)} = \sum_{a,b=1}^2 \w_{ab}^{(\tau)}\; \overline{\vv_a} \; \overline{\vv_b}\trs \ ,
\end{equation}
where $\w^{(\tau)}$ is a $2\times2$-matrix.

\paragraph{Generalizing integrators. }

\begin{figure}[t]
 \centering{
 \includegraphics[width=0.6\textwidth]{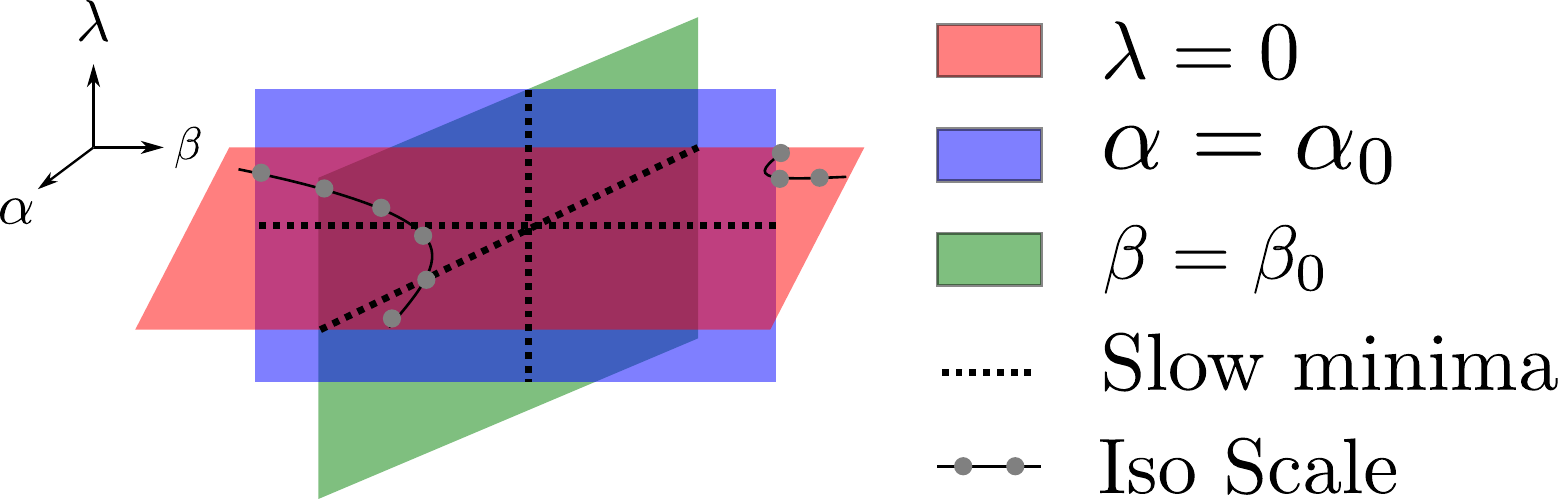}
    \caption{ Illustration of the three $GI$ manifolds in the space of $2\times 2$--matrices with one eigenvalue equal to $\gamma$, the second to $\lambda$, and the remaining two degrees of freedom being labeled $\alpha$ and $\beta$. In one manifold (red), the second eigenvalue is zero, so that all of those matrices are GIs with decay $\gamma$, and any scale $s$. The other two manifolds contain integrators at the particular scale $s^*=\vd\trs\ve$ only, and are of rank $2$. The values of $\alpha_0$ and $\beta_0$ are computed in Appendix \ref{app:integrator_manifolds}, where details on the parametrization used here can also be found. } \label{fig:manifolds}}
\end{figure}

Conditions (\ref{eq:null_loss_conds}) for $\mW$ to be a GI can be turned into conditions over $\mw$, see Appendix \ref{app:lowrank_moments}.  Let us assume that $\mw$ is diagonalized through $\mw = \mP_{\w} \mLambda_{\w} \mP_{\w}^{-1}$ with $\mLambda_{\w}=\text{diag}(\lambda_1, \lambda_2)$, and define $g_i = (\mP_{\w}\trs \sqrt{\mSigma})_{i, 1} (\mP_{\w}^{-1} \sqrt{\mSigma})_{i,2}$. The conditions for $\w$ to define a GI through (\ref{omegatoW}) are: ($\lambda_i=\gamma$ for $i=1$ or 2), ($\sum_i \delta_{\gamma, \lambda_i}g_i = s$) and $g_i=0$ if $\lambda_i\notin \{0,\gamma\}$. Taking into account the constraint $g_1+g_2=\mSigma_{1,2}=\vd\trs\ve$, we find that the set of GIs is spanned by the following three manifolds in the 4-dimensional space of $\w$ matrices, see Appendix \ref{app:integrator_manifolds} for details:

\begin{itemize}
\item The first manifold is of dimension 2, and contains rank-1 integrators $\mW$ at all scales. These weight matrices have one eigenvalue equal to $\gamma$, and the other to $0$ so that one of the $g$ coefficients remains unconstrained:
\begin{equation}\label{rank1}
\mw = \frac{\gamma}{\b-\a}\begin{pmatrix}
\b & -1\\
\a\b &-\a\\
\end{pmatrix} \ ,
\end{equation}
where $(\a, \b)\in \mathbb{R}^2$.
Fixing the scale $s$ to any value different from $\vd\trs\ve$ introduces exactly one relation between $\a$ and $\b$, making the set of rank-1 perfect integrators at scale $s$ a $1$--dimensional manifold, see Appendix \ref{app:integrator_manifolds}.
\item The other two manifolds contain rank-2 integrators, operating at the scale $s^*=\vd\trs\ve$ only. For generic independent encoder and decoder vectors, the scale $s^*=\vd\trs\ve$ is of the order of $n^{-1/2}$ and vanishes in the large size limit. We will discard these solutions, and focus on rank-1 solutions given by (\ref{rank1}) at finite scale $s$ ($\neq \vd\trs\ve$). %Explanations on the structure of minima and convergence towards them are presented in Appendices \ref{app:linear_GI_manifolds} and \ref{app:linear_special_scale}.
\end{itemize}
The structure of the GI manifolds is sketched in Figure \ref{fig:manifolds}.

In Appendix \ref{app:linear_lowrank_hessian}, we compute the gradient and Hessian of the loss in the null-initialization subspace. In the case of fixed encoder and decoder, the convergence towards a GI is generically exponentially fast; the corresponding decay time can be minimized by appropriately choosing the value $s$ of the scale $s$, see Appendix \ref{app:convergence_time}. For some specific choices of the scale $s$, convergence can be much slower and exhibit an algebraic behaviour, see Appendix \ref{app:linear_special_scale}.

\subsection{Initialization with full rank connection matrices}
\label{sec:linear_full_rank_link}

The results above assumed that training started from a null weight matrix, in order to constrain the dynamics of $\mW$ to a very low-dimensional space. Training RNNs on very short epochs ($T=3$) was then sufficient to obtain rank-1 GIs capable of integrating arbitrary long input sequences.

In practice, we observe that initializing the network with a matrix $\mW$ of small spectral norm (instead of being strictly equal to zero) does not change the fact that only one of the eigenvalues of $\mW$ is significantly altered during training, and a GI is obtained as soon as $T\geq 3$. The use of a non-linear optimization scheme such as Adam rather than GD does not change this observation.

To gain insights about this empirical result,
let us consider a perturbation $\bm{\epsilon}=\sum_i \epsilon_i \vu_i \vv_i$, with singular values bounded by 1, around a generalizing integrator of rank 1, $\mW=\sigma\vl\vr\trs$. Under the assumption that the $\vu$ and $\vv$ vectors are drawn randomly on the unit sphere of dimension $n$, their dot products with $\ve$, $\vd$ and each other are realizations of a centered Gaussian distribution of variance $1/n$. We can then consider the image of $\ve$ by our perturbed matrix:

\vspace{-0.4cm}
\begin{equation}
    (\mW+\bm{\epsilon})\ve = (\sigma\vr\trs\ve) \vl + \sum_{i=1}^n \epsilon_i (\vv_i\trs\ve) \vu_i
\end{equation}
\vspace{-0.4cm}

The second term, originating from the perturbation, is a vector whose components are sums of $n$ terms of unfixed signs and magnitudes $1/n$, and is, hence, of the order of $1/\sqrt n$. Accordingly, the dot product of this perturbation vector with $\vd$, which is exactly the perturbation to the first moment $\mu_1$, will be of the order of  $1/\sqrt n$ too. Under similar hypothesis of independance of Gaussian vectors, all moments $\mu_q$ will be perturbed by terms of that same order.

Since unstructured eigenvectors do not contribute to the network output at first order, the gradients with respect to those parameters will also be subleading and this perturbation will remain mostly unchanged during training, in agreement with numerical simulations.

\subsection{Case of multiple channels}
\label{sec:linear2}

We have seen that GD is generally able to train a linear RNN exponentially fast towards a rank--1 single-channel GI with associated eigenvalue $\gamma$ and singular vectors tuned to ensure the correct scale of integration. The state of the  corresponding network is easily interpretable: it is, at all times, proportional to the output integral. Due to the linearity of the network, this result can be straightforwardly extended to the case of $D>1$ integration channels, as we show below.

\paragraph{Interpretation of rank--$1$ solutions in the single channel case.}
We write the rank--$1$ GI as $\mW=\sigma \vl \vr\trs$, where $\vl$ and $\vr$ are normalized to $1$, and $\sigma$ is positive. Since $\mW$ must have $\gamma$ as its eigenvalue, we need $\sigma \vr\trs\vl=\gamma$. Additionally, to ensure that the first non-zero input gives the correct output, we require  that $\sigma (\vd\trs\vl)(\vr\trs\ve)=s\gamma$. It is easy to check that these conditions are sufficient to ensure that the state of the network is
\begin{equation}\label{ch1}
    \vh_t =a\, \overline{y}_t\; \vl\qquad \text{with} \qquad a = \frac 1{\vd\trs\vl} \ ,
    \end{equation}
for all times $t$, which, in turn, ensures perfect integration ($y_t=\overline{y}_t$).
In other words, rank--$1$ GIs rely on a linear, one-dimensional representation of the target integral: the internal state is, at all times, proportional to $\overline{y}_t$.

\paragraph{Representation of integrals with multiple channels.}
The above discussion of the single-channel case generalizes to multiple channels. Through training  a weight matrix $\mW$ of rank $D$ is constructed, which has $(\gamma_1, ..., \gamma_D)$ as its eigenvalues, and singular vectors compatible with the (fixed) encoder and decoder weight vectors. The GI conditions are as follows:
\begin{equation}
\begin{cases}
\forall c\in\llbracket 1, D\rrbracket, &\quad \sigma_c\vr_c\trs\vl_c = \gamma_c\\
\forall c\in\llbracket 1, D\rrbracket, &\quad \sigma_c\,(\vd_c\trs\vl_c)\, (\vr_c\trs\ve_c) = s_c\, \gamma_c\\
\forall (c,c')\in\llbracket 1, D\rrbracket^2, c\neq c', &\quad \vr_c\trs\ve_{c'} = 0\\
\forall (c,c')\in\llbracket 1, D\rrbracket^2, c\neq c', &\quad \vd_c\trs\vl_{c'} = 0\\
\forall (c,c')\in\llbracket 1, D\rrbracket^2, c\neq c', &\quad \vr_c\trs\vl_{c'} = 0\\
\end{cases}
\end{equation}
The two first conditions are exactly the same as in the single channel case, while the last three ensure that the modes of the weight matrix coding for the different integration channels $c$ do not interfere, and can independently update the values of their outputs to match the targets $\overline{y}_{c,t}$.

Assuming these conditions are satisfied, the network state is at time $t$ equal to
\begin{equation}
\vh_t=\sum_{c=1}^D a_c\, \overline{y}_{c,t}\, \vl_c \ ,
\end{equation}
where the $a_c$'s are structural coefficients, which generalizes expression (\ref{ch1}) to the case $D\ge 2$. The state of any neuron $i$ is therefore a linear combination of the $D$ integrals across the multiple channels. Multiplexing is here possible as long as $D\le n$, and encoders and decoders each form free families of $\mathbb{R}^n$.

%%%%
%
% Non linear D=1
%
%%%%

\section{Non-linear activation: case of a single channel}
\label{sec:single_channel}

We now turn to the case of non-linear activation. The computation of the averaged loss is not analytically feasible any longer. However, by investigating RNNs trained with Gradient Descent on the mean square error~(\ref{loss1}) computed on batches of inputs, hereafter referred to as batch--SGD, we have identified structural and dynamical properties, from which sufficient conditions for generalization can be constructed.

\subsection{Empirical study of neural representations in a ReLU network}

We start by considering the case of the ReLU activation, where $f=\R{\cdot}=\max(\cdot,0)$ is a non-linear  component-wise operator.
The simple encoding (\ref{ch1}) adopted by linear-activation networks relied on the fact that the activity of each neuron could change sign with $\overline{y}_t$. This is not possible with ReLU activation anymore since activities are forced to remain non-negative, and a novel encoding is obtained after training of the RNNs that we expose below.

\begin{figure}
\centering
    \includegraphics[width=\textwidth]{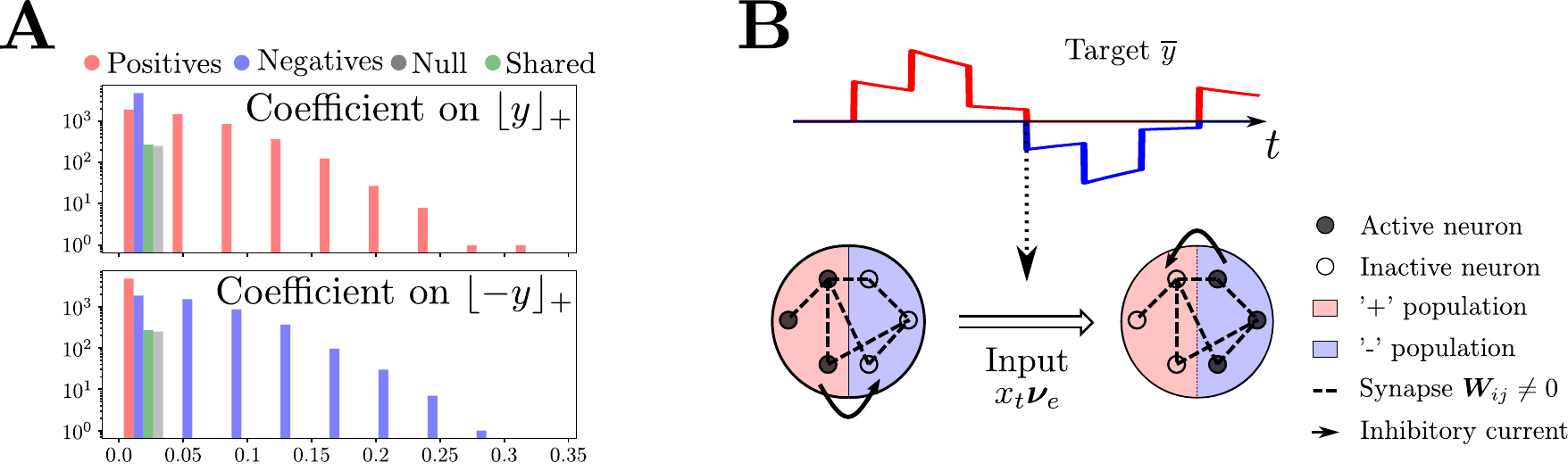}
    \caption{ Internal encoding of the integral $y_t$ by a single-channel ReLU network using two populations. \textbf{A}: Experimentally observed distributions of the components of $\mL_{\pm}$, determined by fitting the activity of each neuron with (\ref{eq:relu_coding}). Results are aggregated across 10 realizations of batch--SGD training $n=1000$, $s=1$. \textbf{B}: Illustration of the activity shift from the $+$ to the $-$ population at arrival of an input that changes the sign of the target. Mutual inhibition between the two sub-networks guarantees only one can be active at a given time, and an external input is required to perform the shift.   \label{fig:bipopulation_coding}}
\end{figure}

\paragraph{Behavior of neuron activities.}
 Based on  numerical simulations reported in Figure \ref{fig:bipopulation_coding}A, we argue that the population activity in ReLU networks depends on two vectors, referred to as  $\mL_{+}$ and $\mL_-$, with non-negative components and dot products with $\vd$ equal to, respectively, $+1$ and $-1$. More precisely, these vectors determine how the neural activities vary with the integral $\overline{y}_t$, depending on its sign:
\begin{equation}
\label{eq:relu_coding}
  \vh_t = \R{\overline{y}_t} \,\mL_+ + \R{-\overline{y}_t}\, \mL_- \ .
\end{equation}
Hence, in the space of possible internal states $\mathbb{R}_+^n$, the state $\vh$ of the RNN lies in the union of the two half lines along $\mL_+$ and $\mL_-$, a 1-dimensional piecewise linear manifold whose geometry is imposed by the non-linear activation $\R{\cdot}$.

The $n$ components of $\mL_+,\mL_-$ define \textit{a priori} four sub--populations: if $(L_+)_i> 0$ and $(L_-)_i> 0$ neuron $i$ is active at all times $t$ ("shared" population); if $(L_+)_i> 0$ and $(L_-)_i= 0$ (respectively, $(L_+)_i= 0$ and $(L_-)_i> 0$), the neuron is only active when the integral is positive (resp. negative), defining the "+" (resp. "-") population; if $(L_+)_i= (L_-)_i= 0$ , the neuron is never active and belongs to the "null" population. In numerical experiments, the shared and null populations account for a small fraction of the neurons (around $5\%$, see Figure~\ref{fig:bipopulation_coding}A) when training is performed using batch--SGD; in addition, shared neurons never have strong activities and their contributions to the output integral seem irrelevant. We introduce in equation (\ref{eq:proxy_loss_relu}) a new loss function, which allows for training of perfect integrators that do not exhibit any shared or null neurons.

\begin{figure}
\centering
    \includegraphics[width=\textwidth]{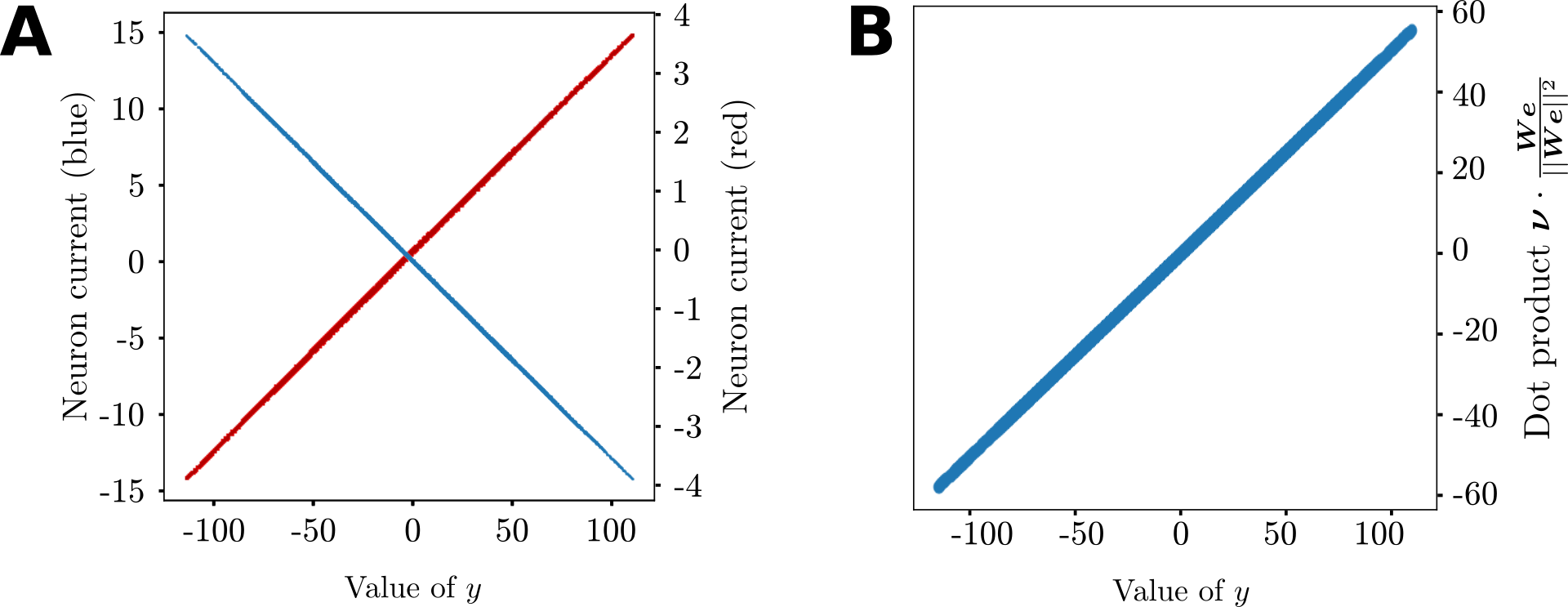}
    \caption{ Behavior of currents in a ReLU network trained using the batch--SGD loss, $s=2$, $\gamma=0.995$. \textbf{A}: Parametric plot of the currents $(\vnu_t)_i$ incoming on two representative neurons $i$ (red, blue) vs. target integral $y_t$ across time $t$. We observe a linear relation, with a slope that varies both in sign and magnitude from neuron to neuron. \textbf{B}: Normalized dot product between the vector of currents $\vnu$ and the image of the encoder $\mW\ve$ vs. value of the integral, illustrating eqns. (\ref{curr_relu}) and (\ref{eq:LWe}). \label{fig:relu_currents_two_panels}}
\end{figure}

\paragraph{Behavior of neuron currents.}
Numerical experiments furthermore indicate that the dependence of the current $\vnu_t$ (\ref{defcurr}) on the integral $y_t$ is simpler than the one shown by the activity $\vh_t$. We observe that the current vector is proportional to the integral,
\begin{equation}\label{curr_relu}
    \vnu_t = y_t\, \mL,
\end{equation}
where the components $L_i$ of the vector $\mL$ vary from neuron to neuron, both in amplitude and in sign, see  Figure \ref{fig:relu_currents_two_panels}A.

The representation of the integral based on two non-overlapping populations reported above may be seen as a straightforward consequence of the linear encoding at the level of pre-activation currents expressed by (\ref{curr_relu}):
\begin{equation}\label{Lpm_definition}
    \vh_t = \R{\vnu_t} = \R{y_t\; \mL} = \R{y_t}\;\R{\mL} + \R{-y_t}\;\R{-\mL },
\end{equation}
from which we deduce that the population vectors $\mL_+$ and $\mL_-$ defined in (\ref{eq:relu_coding}) are equal to, respectively, $\R{\mL}$ and $\R{-\mL}$. In other words, neurons $i$ encode positive or negative values of the integral depending on the signs of the components $L_i$.

Hence, while the neural state $\vh _t = f(\vnu_t)$ of a ReLU RNN is not proportional to the integral value, see (\ref{eq:relu_coding}),
 as was the case for linear RNNs in Section \ref{sec:linear_full_rank_link}, proportionality is recovered at the level of the pre-activation currents $\vnu_t$. We will see below that the linearity of the currents with respect to the integrals extends to the case of multi-channel integrators.

\subsection{Theoretical analysis of the ReLU integrators}
\label{sec:relu_D1_theory}

We now explain the origin of the linear relationship between current and integral values (\ref{curr_relu}), and how the vector $\mL$ defining the current direction is related to the connection matrix $\mW$, the encoder $\ve$, and the parameters $s,\gamma$.

\paragraph{Sufficient conditions for  integration.}
Let us first consider the network at time $t=0$, with all activities set to zero ($\vh_0=\bm{0}$). As the first input $x_1$ is read by the encoder, the current vector at time $t=1$ takes value
\begin{equation}
    \vnu_1 = \mW(\bm{0} + x_1 \, \ve) = x_1 \;\mW\ve = \frac{\bar{y}_1}{s \gamma}\;\mW\ve \ .
\end{equation}
The above equality agrees with the linear relationship (\ref{curr_relu}) provided we have
\begin{equation}
\label{eq:LWe}
 \mL   = \frac 1{s \gamma}\;  \mW\ve  \ .
\end{equation}
This identity is in excellent agreement with numerical findings, as shown in Figure \ref{fig:relu_currents_two_panels}B.

\begin{figure}
\centering
    \includegraphics[width=0.8\textwidth]{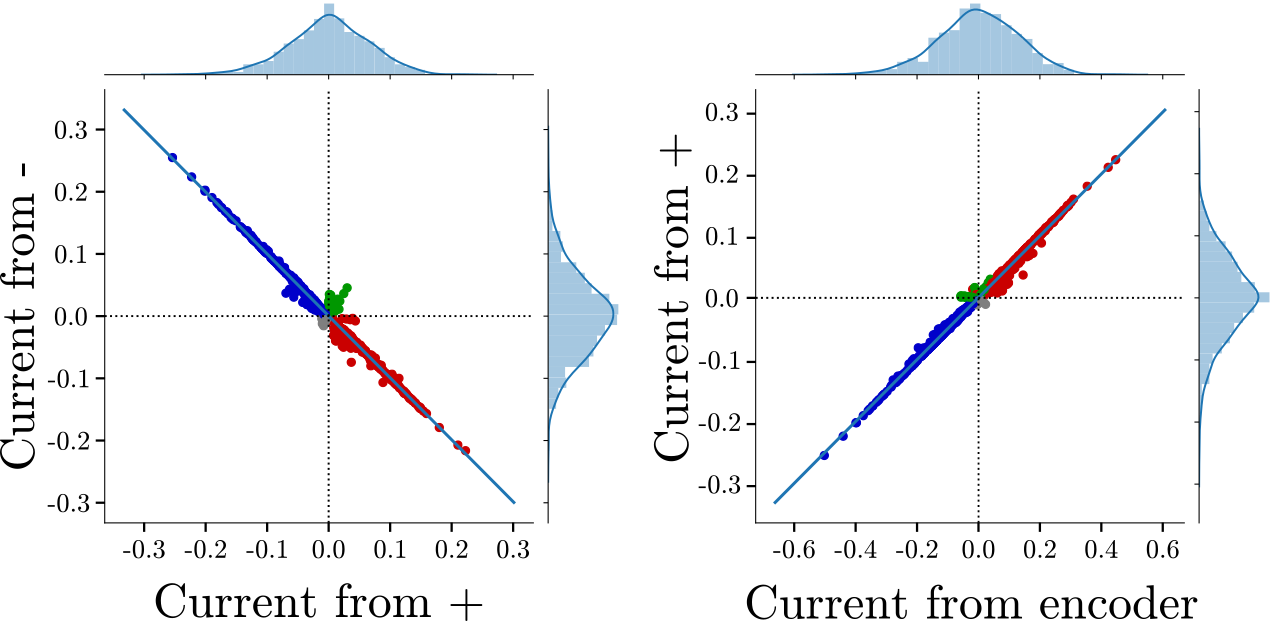}
    \caption{ Contributions to the currents in a ReLU integrator trained with batch--SGD. Left: scatter plot of  $\mW \mL_-$ vs. $\mW \mL_+$. Right: $\mW\mL_+$ vs. $\mW\ve$. Colors refer to the neural populations, see Figure~\ref{fig:bipopulation_coding}A.  For both panels we show on the sides the histograms of current components. Results were obtained with $T=10$, $\gamma=0.995$, $s=2$, $n=1000$. Numerical findings confirm that $\mW\mL_+ = -\mW\mL_-$ and $\mW\ve = s\;  \mW\mL_+$. \label{fig:current_patterns}}
\end{figure}

We now assume that the current linearly expresses the target integral $\bar y_t$ at time $t$, and look for sufficient conditions for  relationship (\ref{curr_relu}) to hold at time $t+1$ after the new input $x_{t+1}$ is received by the network. The current at time $t+1$ reads
\begin{equation}\label{proof_linear_encoding}
\begin{split}
    \vnu_{t+1} &= \mW (\vh_{t}  + x_{t+1}\, \ve) = \mW(\R{\vnu_{t}}  + x_{t+1}\, \ve) \\
    &= \mW\left (\R{\bar{y}_t \,  \mL}  + x_{t+1}\, \ve \right) = \mW(\left\R{\bar{y}_t}\, \mL_+ + \R{-\bar{y}_t}\, \mL_- \right) + x_{t+1} \, \mW \ve \\
    &= \R{\bar{y}_t}\, \mW\mL_+ +  \R{-\bar{y}_t}\,\mW\mL_- + x_{t+1}\, \mW\ve \ ,
\end{split}
\end{equation}
and should match
\begin{equation}
    \vnu_{t+1} = \frac{\bar y_{t+1}}{s\gamma} \, \mW\ve = \big(\frac{\bar y_{t}}s+x_{t+1}\big) \, \mW\ve
\end{equation}
according to (\ref{curr_relu}) and (\ref{eq:LWe}). We deduce that $\mW\mL_+$ and $\mW\mL_-$ have to be aligned along  $\mW\ve$, see (\ref{eq:LWe}). Furthermore, based on the identity $y=\R{y}-\R{-y}$, we readily obtain
that
\begin{equation}\label{eq:currents_relation}
    \mW \mL_+=-\mW\mL_-=s^{-1}\, \mW\ve \ .
\end{equation}
These relations are in very good agreement with numerics, see Figure~\ref{fig:current_patterns}.

\paragraph{Proxy loss for integration by a network of ReLU units.}
\label{sec:relu_proxy}
Conditions (\ref{eq:LWe},\ref{eq:currents_relation}) as well as the relations between $\mL, \mL_+,\mL_-$ ensure perfect generalizing integration. They  can be summarized into the set of four equalities
\begin{equation}\label{condloss}
    \begin{cases}
     \vd\trs \R{\pm \mW \ve} &= \pm s\gamma \\
     \mW \R{\pm \mW \ve} &= \pm \gamma \mW \ve \\
    \end{cases}
\end{equation}
linking the matrix of connections, the encoder and decoder vectors, as well as the scale and decay parameters.

We now introduce a proxy loss for $\mW$, whose global minimum is achieved when  conditions (\ref{condloss}) above are fulfilled,
\begin{equation}
\label{eq:proxy_loss_relu}
    \mathcal{L}^{proxy} = \sum _{z=\pm1} (\vd\trs \R{z \mW \ve} - \, z s \gamma)^2+ \sum _{z =\pm1} |\mW \R{z \mW \ve} - \, z  \gamma \mW \ve|^2\ .
\end{equation}
Experimentally, training on this proxy loss is extremely effective and as expected leads to perfect integrators satisfying the relations between currents shown in Figure \ref{fig:current_patterns}. Similarly to the linear case, if the encoder and decoder are fixed during training, the convergence time of GD is strongly dependent on $s$ with a preferred scale around $|\ve||\vd|$, see study of dynamics of learning with $\mathcal{L}^{proxy}$ in Appendix \ref{app:relu_proxy_gradients},

While the batch-SGD loss is by definition based on actual computation of the network output for sample input sequences, the proxy loss imposes strict conditions on the dynamical behavior of the network that, in turn, ensure that the batch-SGD loss will be zero. While there is no {\em a priori} reason to believe that all global minima of (\ref{loss1}) are global minima of (\ref{eq:proxy_loss_relu}), we empirically observed that the solutions $\mW$ found by minimizing the batch--SGD seemed to also be approximate minima of the proxy loss (see Figure \ref{fig:current_patterns} for the ReLU case).

\paragraph{Properties of the connection matrix.} Training integrators with either batch--SGD or the proxy loss yields solutions with one dominant singular value, of the form
\begin{equation*}
    \mW\simeq\sigma\,\vl \, \vr\trs.
\end{equation*}
We report some properties of these solutions in Appendix \ref{app:rank_1_relu}. In particular, the singular value $\sigma$ is, in the case of fixed encoder and decoder with unit norms, bounded from below by $2 \max(1, s)$, where $s$ is the scale. In practice, except for scales close to $1$, this lower bound seems to be tight, {\em i.e.} $\sigma = 2\max(1,s)$, see Appendix \ref{app:rank_1_relu}, Figure \ref{fig:understand_relu_solution}. We interpret this saturation as a manifestation of the conjecture by \citep{ImplicitRegularization} that gradient descent implicitly favors solutions with small matrix norm, as rank--1 matrices have a Frobenius norm equal to their singular value.

\subsection{Case of generic non-linear activation.} \label{sec43}
We now turn to the generic case of  non linear activation function $f$. To do so, we show how the idea of proxy loss developed in the ReLU case can be naturally extended to any $f$.

\paragraph{Generic proxy loss.} We start by writing, for an arbitrary activation function $f$, the dynamical equation for the current, rather than for the activity state,
\begin{equation}
    \vnu_{t+1} = \mW\big(f(\vnu_t) +x_{t+1} \ve\big).
\end{equation}
At the first time-step, since $\vh_{-1}=\bm{0}$ the current  will be equal to $\vnu_0=x_0 \mW\ve = y_0/(s\gamma) \mW\ve$. The error will thus vanish if and only if, for all $y$ in the range of values of the target integral,
\begin{equation}\label{cond1}
     \vd\trs f\bigg(\frac{y}{s\gamma}\, \mW\ve\bigg) = y.
\end{equation}
These relations generalize the first two conditions in (\ref{condloss}) for ReLU activation.
%, and demand that the line \vnu_e$ becomes "adapted" to the decoder.
Furthermore, imposing that $\mW\ve$ is an `eigenvector' of the non-linear operator $\mW f(\cdot)$ with eigenvalue $\gamma$, \textit{i.e.}
\begin{equation}\label{cond2}
        \mW \cdot f\bigg(\frac{y}{s\gamma}\, \mW \ve\bigg)= \frac{y}{s} \, \mW\ve,
\end{equation}
for any $y$, will force the current to remain at any time aligned along  $\mW\ve$. A simple inductive proof similar to (\ref{proof_linear_encoding}) shows that in these idealized conditions the coordinate along that line will evolve proportionally to the output, similarly to eqn. (\ref{curr_relu}). Combined with the condition derived for the first input, this is enough to guarantee perfectly generalizing integration.

For arbitrary $f$, conditions (\ref{cond1}) and (\ref{cond2}) can generally not be exactly satisfied for $y$ varying over a continuous domain, {\em i.e.} for an infinite number of values of $y$. However, these conditions can be fulfilled for a discrete and finite subset, which will provide sufficient accuracy for good integration in practice, and we observe that the error on the integral of a time series of $T$ inputs to scale as $\epsilon\sim n^{-1/2}$, irrespectively of $T$ (as long as the integral values remains below $y_{max}$).

Based on these considerations, we propose a proxy loss for integration of a single scalar signal using a RNN with arbitrary non-linearity:
% \begin{equation}\label{eq:general_proxy_D1}
%     \mathcal{L}^{proxy, f, D=1}(\mW) = \int_{y \in A} \left[\vd\trs f(\alpha \mW\ve) - s \,\gamma\, \alpha\right]^2 + \left[ \mW\cdot f(\alpha\, \mW\ve) -  \alpha\,\gamma\, \mW\ve\right]^2.
% \end{equation}
\begin{equation}\label{eq:general_proxy_D1}
    \mathcal{L}^{proxy, f, D=1}(\mW) = \int_{z \in Z} \left[\vd\trs f(z \mW\ve) - s \,\gamma\, z\right]^2 + \left[ \mW\cdot f(z\, \mW\ve) -  z\,\gamma\, \mW\ve\right]^2.
\end{equation}

This integral can be estimated via Monte-Carlo, and the choice of $Z=[-z_{max}, z_{max}]$ will restrict the maximum value $y_{max}=s\, \gamma \, z_{max}$ of $y$ that can be represented through our network. It is still possible to obtain generalization to infinite number of integration steps, but the choice of $\gamma$ has to be tuned so that the integral never exceeds the range the network was trained for.

\begin{figure}
\centering
    \includegraphics[width=\textwidth]{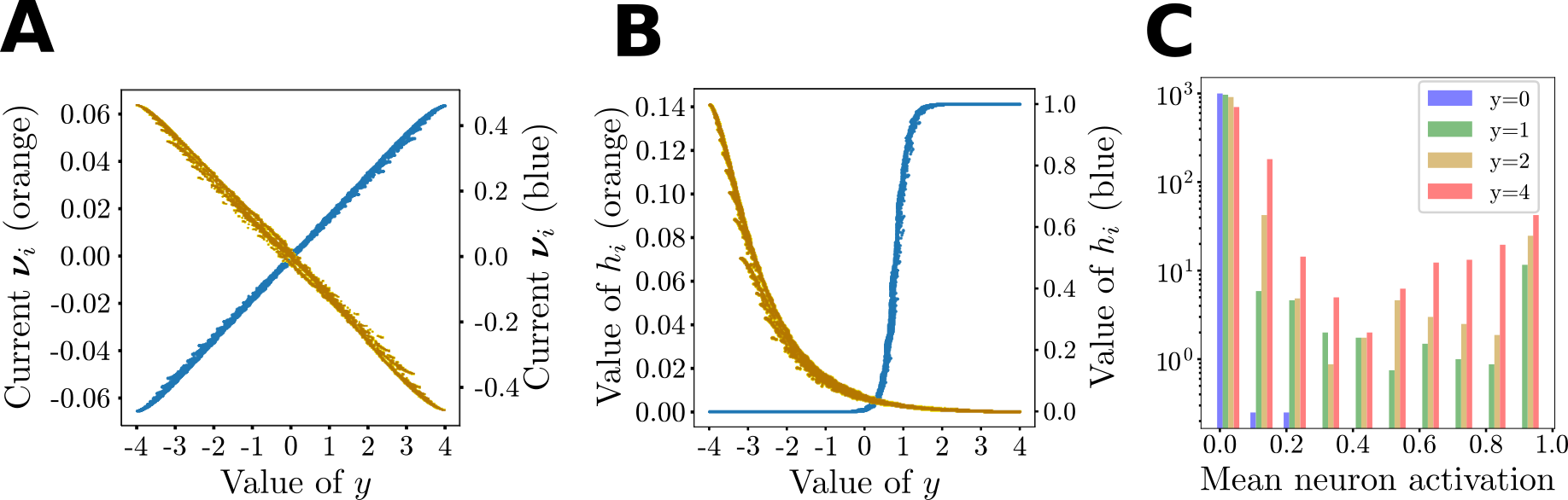}
    % there is V0 (original) up to V3, V2 is my favorite but you can change it
    \caption{\textbf{A}: Value of the pre-activation current $\nu_i$ as a function of the integral for two representative neurons. \textbf{B}: Activity-integral characteristic curve for the neurons of panel \textbf{A}. One of them (blue,  right scale) saturates for low enough values of $y_t$, while the other (orange, left scale) never saturates. \textbf{C}: Histogram of the mean activity of neurons for different values of the integrals, aggregated across $8$ realizations of the training on the proxy loss (\ref{eq:general_proxy_D1}). The range of integral values $[-4, 4]$ was divided in $100$ bins to select the time-steps in the test sequences that corresponded to the values of $y$ indicated in the legend. As the value of the integral increases, more neurons get strongly activated, and eventually saturate. The same evolution could be observed for integrals $y_t$ decreasing below the zero value. Those networks were trained using the batch--SGD loss, $\gamma=0.8$, $s=1$, $n=1000$, and the same results are found using the proxy loss.}
    \label{fig:sigmoid_D1}
\end{figure}

\paragraph{Application to sigmoidal activation.}
We tested this new loss with a sigmoidal activation function\footnote{ The choice of the slope and bias, here $50$ and $0.1$ respectively, is not critical to the results. We chose the slope so that the transition from $0$ to $1$ of the sigmoid happens on a scale of $1/50$, close to the expected magnitude of the currents $n^{-1/2}\simeq 1/30$ for $n=1000$. The bias was then chosen so that $x=0$ is not in the linear portion of the sigmoid, nor in a fully saturated portion to avoid the null weight-matrix $\mW^{(0)}=\bm{0}$ to be a fixed point of the learning dynamics.} \begin{equation*}
    f:x\rightarrow\frac{1}{1+\exp^{-50(x-0.1)}}.
\end{equation*}
Trained with a decay $\gamma=0.8$, scale $s=1$, $Z=[-5,5]$~\footnote{ For $\gamma=0.8$, $s=1$, and inputs of magnitude bounded by $1$, the integral evolves in $[-4, 4]$ as $y_{max}$ is solution of $y_{max} = \gamma(y_{max} + s)$, hence $z_{max} = y_{max}/ (s\gamma) = 5$. In practice, to observe the regimes $|y|\simeq y_{max}$ more easily, we test the network using sequences alternating between bursts of $\pm 1$ inputs and long periods with no external input, see Figure \ref{fig:sigmoid_D2}.}, those networks converge to a solution with a single dominant singular value and manage to integrate signals of arbitrary length, despite their inability to generalize to larger values of the integral. We observe that some neurons in the network exhibit a saturated behaviour when the integral is above (resp. below) a neuron-specific threshold $\theta_i$, while other neurons never reach that saturation. This results in a behavior where, during monotonous evolution of the integral starting from $0$, an increasing number of neurons get activated to support the integral, see Figure \ref{fig:sigmoid_D1}. While these networks have a very different phenomenology from the ReLU ones in state space, the integration is still performed through linear currents. We also confirmed that sigmoidal networks could be trained on the batch-SGD loss, yielding integrators with a single dominant singular value; training with $\gamma$ too close to 1 results in poor performance, suggesting that the issues of generalization to large values of $y$ is not entirely due to the choice of proxy loss, but could hint at intrinsic limitations of the network, related to the activation function.

The proxy loss (\ref{eq:general_proxy_D1}) will be extended below to the general case $D>1$. It should be noted that  all non-linear integrators need not be absolute minima of the proxy loss and follow the linear current representation. We only show here that it is one possible representation scheme, which can be adapted to any non-linearity and could therefore help bridge the gap between idealized ReLU activation and more complex examples, e.g. inspired from real neurons.

%%%%
%
% Non linear D >= 2
%
%%%%

\section{Non-linear activation: case of multiple channels}
\label{sec:multi_channel}

We now consider the case of a multiplexed integrator with $D$ input-output channels, performing $D$ integrals in parallel. In practice, numerical experiments were carried out  for $D=2,3,4$.

\paragraph{Batch and proxy losses for multiple integrators.}

To train our RNN to carry out multiple integrations, we followed two different strategies. First, we used the batch loss defined in (\ref{loss1}) from a set of input data, combined with a learning algorithm, {\em e.g.} SGD.

Second, drawing our inspiration from the detailed analysis of the single-channel case studied in the previous section, we introduced an extension of  the proxy loss (\ref{eq:general_proxy_D1}) to an arbitrary number $D>1$ of input signals,
% \begin{equation}
% \begin{split}
% \mathcal{L}^{proxy, f, D}(\mW) = \int_{y_1 \in A_1} \dots \int_{y_D \in A_D}& \bigg\{ \sum_c \left[\vd_c\trs f\big(\sum_c y_c \mW\ve_c\big) - s_c\, \gamma_c\, y_c\right]^2\\ & \hspace{-1cm} + \left[\mW\cdot f\big(\sum_c y_c \mW\ve_c \big) - \sum_c  \gamma_c\, y_c \mW\ve_c \right]^2 \bigg\} \ ,
% \end{split}
%     \label{eq:general_proxy}
% \end{equation}

\begin{equation}
\begin{split}
\mathcal{L}^{proxy, f, D}(\mW) = \int_{z_1 \in Z_1} \dots \int_{z_D \in Z_D}& \bigg\{ \sum_c \left[\vd_c\trs f\big(\sum_c z_c \mW\ve_c\big) - s_c\, \gamma_c\, z_c\right]^2\\ & \hspace{-1cm} + \left[\mW\cdot f\big(\sum_c z_c \mW\ve_c \big) - \sum_c  \gamma_c\, z_c \mW\ve_c \right]^2 \bigg\} \ ,
\end{split}
    \label{eq:general_proxy}
\end{equation}
where the integral runs overs the $D$-dimensional range of values of the integrals, $Z_1\times Z_2\times ...\times Z_D$. As we shall see below, training with this loss allowed us to obtain networks with arbitrary non-linearity that represent the integral values linearly in the space of currents, as we shall see below. Note that different activation functions, varying from neuron to neuron could be also considered, {\em e.g.} through the introduction of a distribution of thresholds for the sigmoidal function.

\begin{figure}
\centering
    \includegraphics[width=\textwidth]{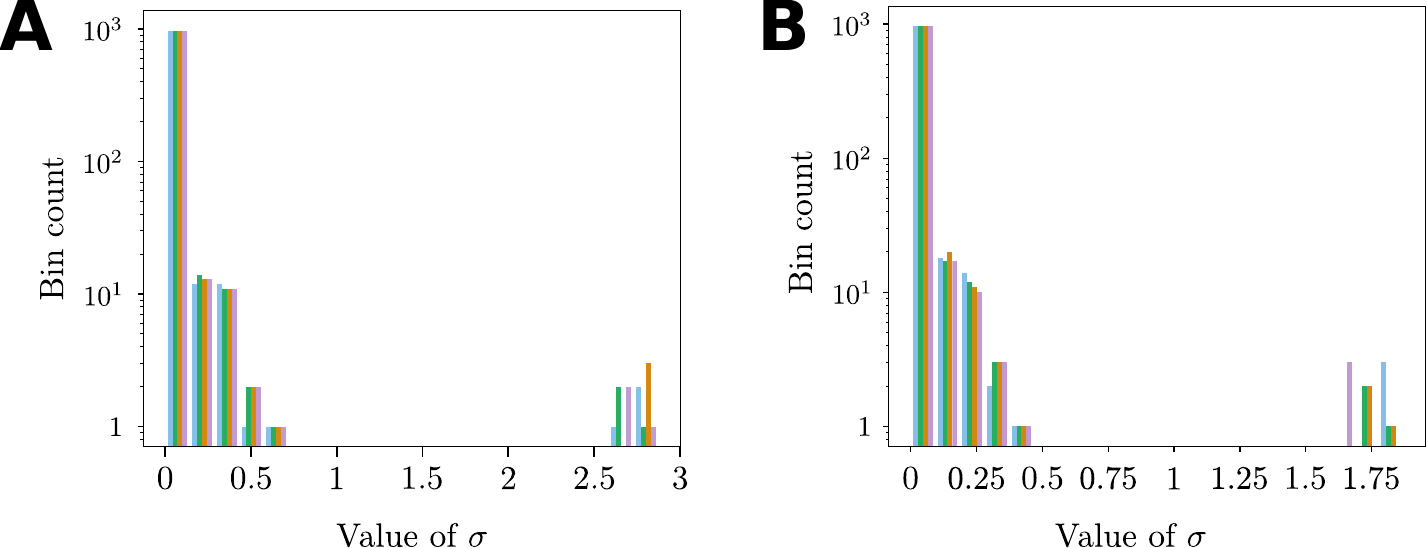}
    \caption{ Histograms of the singular values of $\mW$ in a ReLU (\textbf{A}) and sigmoidal (\textbf{B}) network across 4 realizations (one colour each) of batch--SGD with $D=3$, $T=10$, $n=1000$. The ReLU networks were trained with $\gamma_1=\gamma_2=.995$, $\gamma_3=.992$, while the sigmoidal ones were trained with $\gamma_1=\gamma_2=.8$, $\gamma_3=.75$. In both cases, a \textit{bulk} of eigenvalues are found close to $0$, while exactly $3$ of them become substantially larger. A fair amount of variability can be observed in the exact value of those large eigenvalues, even using the same values of the decays.}
    \label{fig:svd_histograms}
\end{figure}

\paragraph{Characterization of currents for ReLU networks.}
We start with the ReLU case. As in the linear case, training ReLU networks with Stochastic Gradient Descent of the batch loss yields networks that perform multiple integrations with excellent accuracy. Inspection of the connection matrices $\mW$ reveals that they have $D$ dominant singular values,
%and strongly overlapping singular vectors $\vl$,
as illustrated in Figure \ref{fig:svd_histograms}A for $D=3$ channels. Such a spectral structure, consisting of a large number of "bulk" values and a small number of "outliers" that perform a computational task is reminiscent of the setting investigated in \citep{schuessler_dynamics_2020, schuessler_interplay_2020}.

\begin{figure}
\centering
    \includegraphics[width=.85\textwidth]{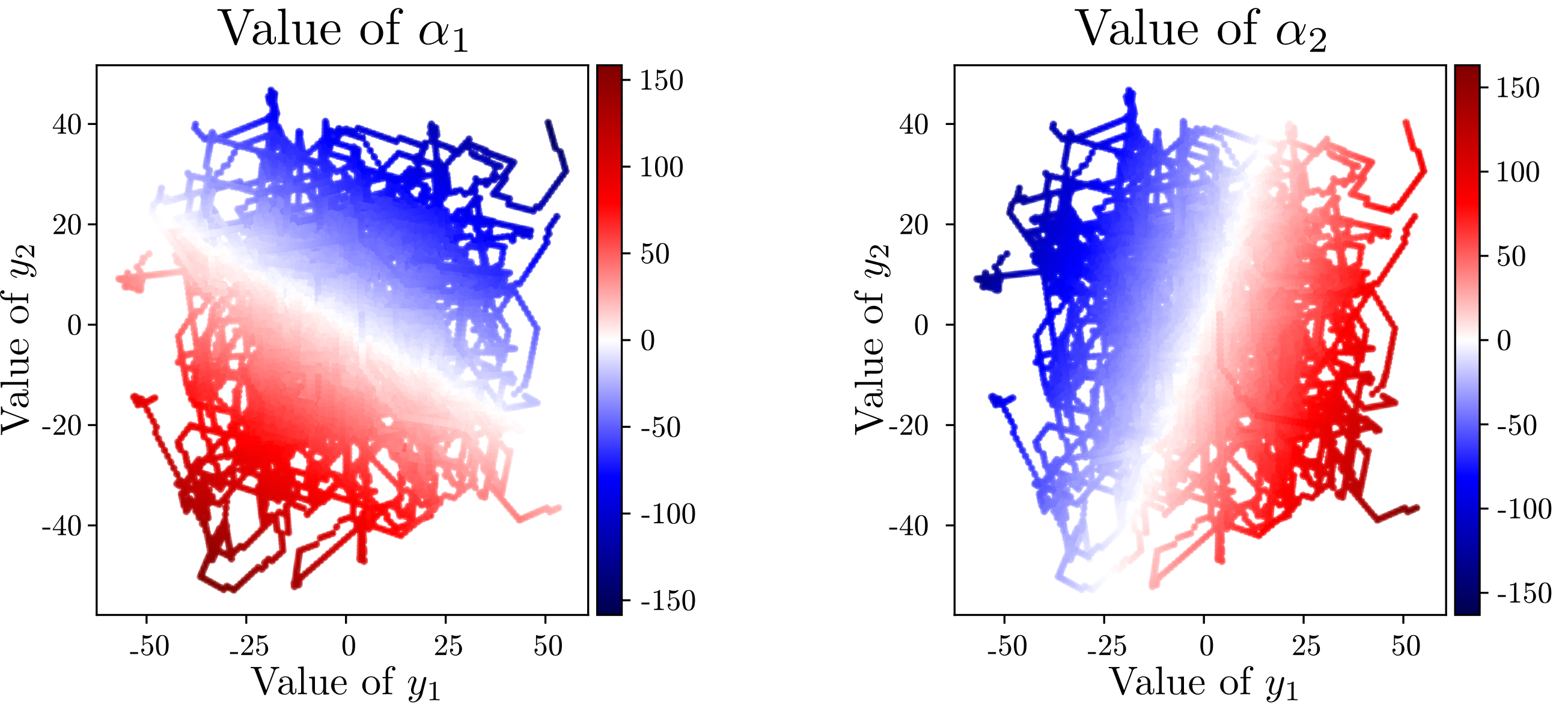}
    \caption{ Value of the coordinates $\alpha_1$ and $\alpha_2$ in the current manifold as a function of the value of the target outputs $\bar{\vy}$. Both coordinates depend linearly on the value of the two integrals $(y_1, y_2)$, so that the position in the current manifolds is a linear representation of the integrals. The points were aggregated across $256$ trajectories of length $T^{test}=200$, for networks trained using batch--SGD on the mean square error (\ref{loss1}) with training epochs duration $T^{train}=10$, $\gamma_1=0.995$, $\gamma_2=0.992$. }
    \label{fig:bi_channel_map}
\end{figure}

The $D$ corresponding left eigenvectors $\vl_c$ of the $\mW$ matrix define a $D$--dimensional linear manifold for the current vector $\vnu_t$,
\begin{equation}\label{rel1}
    \vnu_t \simeq  \sum_{c=1}^D \alpha_{c,t} \,\vl_c , \ (\a_{1,t}, .., \a_{D,t})  \in\mathbb{R}^D\ ,
\end{equation}
while the activity state $\vh_t$ of the network lives on a non-linear version of this manifold, shaped by the ReLU activation function:
\begin{equation}\label{rel1b}
    \vh_t =\R{ \vnu_t}\ .
\end{equation}
Investigating the relation between the $\alpha$ coordinates in the current manifold and the values of the different integrals $\overline{\vy}$, we empirically find that they are related by a linear mapping. More precisely, there exists a $D\times D$--matrix $\mR$ such that the coordinates $\boldsymbol\alpha_t$ along the current-manifold can be written at all times as:
\begin{equation}\label{rel2}
    \alpha_{c,t}= \sum_{c'=1}^D R_{c,c'} \;\overline{y}_{c',t}\ .
\end{equation}
In Figure \ref{fig:bi_channel_map}, we illustrate this mapping in the $D=2$ case. The methodology adopted is the following. While the network is performing integration, at each time-step $t$, we infer the $\a_{c,t}$ coordinates from the values of the currents through (\ref{rel1}). The panels of Figure \ref{fig:bi_channel_map}  show the coordinate $\alpha_c$ (left: $c=1$; right: $c=2$), see color code in the figure, as a function of the two integrals  $\overline{y}_1, \overline{y}_2$. Aggregating those results across a large number of long trajectories, we find that the value of the currents as a function of the targets is independent of the exact input sequence and linearly depends on the value of the integrals. Hence, the linear dependence of the current on the integrals, empirically found for $D=1$ in (\ref{curr_relu}), also holds in the multi-channel case.

 We emphasize that the presence of a bulk of small, but not negligible, singular values of $\mW$ (in addition to the $D$ dominant ones) is not in contradiction with the fact that the current lives in a $D$--dimensional manifold. The corresponding singular vectors may be orthogonal to the encoders, and therefore never contribute to the internal state.  To illustrate this point, we provide a quantitative evaluation of the distance between the currents $\vnu_t$ and the $D$--dimensional vector space $\mathcal{D}$ spanned by the $D$ largest singular vectors $\vl_c$ on the right hand side of eqn~(\ref{rel1}) as follows. After collecting the currents $\vnu$ at all time-steps during $128$ trajectories of duration $T=200$, we compute the projection $\vnu^{\parallel}_t$ of those currents on  $\mathcal{D}$ using least-squares, and the orthogonal projection, $\vnu^\bot_t$. The ratio of their norms
\begin{equation}\label{ratio_def}
r= \frac{\left \langle |\vnu^\bot_t| \right \rangle_t}{\left \langle |\vnu^{\parallel}_t| \right \rangle_t} \ ,
\end{equation}
where $\langle \cdot\rangle_t$ denotes the average over time, estimates how much of the current lies out of the $D$-dimensional manifold. Results for the ratios are reported in the first line of Table \ref{fig:orth_norm_ratio} for networks obtained from the batch and the proxy losses, and are very small, $r< 0.5$. These values are significantly smaller than what would be expected by chance in a null model in which all directions in the $n$-dimensional space of currents would be equally significant,
\begin{equation}
    r_{null} = \sqrt{\frac nD-1} \ ,
\end{equation}
whose value is larger than 20 for $n=1000$ and $D=1,2$.

% \begin{table}
% \begin{center}
%   \begin{tabular}{ c | c | c | c | c |}
%     \, & D=1, proxy & D=2, proxy & D=1, batch & D=2, batch \\ \hline
%     ReLU & $\scriptstyle 1.63\,10^{-2} \pm 9.22\,10^{-4}$ & $\scriptstyle4.9\,10^{-2} \pm 4.87\,10^{-3} $ & $\scriptstyle 1.31\cdot10^{-2} \pm 1.69\,10^{-3}$ & $\scriptstyle 1.25\,10^{-2} \pm 1.38\,10^{-3}$ \\ \hline
%     Sigmoid & $\scriptstyle 3.22\cdot10^{-2} \pm 2.44\cdot10^{-3}$ & $\scriptstyle1.13\cdot10^{-1} \pm 1.36\cdot10^{-2}$ & $\scriptstyle5.62\cdot10^{-2} \pm 1.45\cdot10^{-2}$ & $\scriptstyle3.34\cdot10^{-1} \pm 1.90\cdot10^{-2}$ \\
%     \hline
%   \end{tabular}
%   \caption{Average ratios $r$ of the projections of the current outside and inside the best $D$--dimensional subspace, see eqn. (\ref{ratio_def}), for different activation functions and values of $D$, and $n=1000$. Error bars were estimated from $8$ realisations of the training in the same conditions.}\label{fig:orth_norm_ratio}
% \end{center}
% \end{table}

% \begin{table}
% \begin{center}

%     \, & D=1, proxy & D=2, proxy & D=1, batch & D=2, batch \\ \hline
%     ReLU & $\scriptstyle 1.63\,10^{-2} \pm 9.22\,10^{-4}$ & $\scriptstyle4.9\,10^{-2} \pm 4.87\,10^{-3}

% \end{table}

\begin{table}
\begin{center}

  \begin{tabular}{ c | c | c | c | c |c | c |}
     \cline{2-7}
    \multicolumn{1}{c}{$\times 10^{-2}$} & \multicolumn{2}{|c|}{Batch} & \multicolumn{4}{|c|}{Proxy}\\
    \cline{2-7}
    \, & D=1 & D=2 & D=1 & D=2 & D=3 & D=5 \\ \hline
    \multicolumn{1}{|c|}{ReLU} & $1.3 \pm 0.2$ & $1.3 \pm 0.1$ & $1.63 \pm 0.1$ & $4.9 \pm 0.5$ & $3.3 \pm 0.5$ & $2.5 \pm 0.1$ \\ \hline
    \multicolumn{1}{|c|}{Sigmoid} & $5.6 \pm 1.5$ & $33.4 \pm 1.9$ & $3.22 \pm 0.2$ & $11.3 \pm 1.4$ & $10.0 \pm 0.4$ & $5.9 \pm 0.3$ \\ \hline
  \end{tabular}
  \caption{ Average ratios $r$ of the projections of the current outside and inside the best $D$--dimensional subspace, see eqn. (\ref{ratio_def}), for different activation functions and values of $D$, and $n=1000$. Errors were estimated from $8$ realisations of the training in the same conditions, and all values reported in the table are $10^2 \times r$ for readability.}\label{fig:orth_norm_ratio}
\end{center}
\end{table}

\begin{figure}
\centering
    \includegraphics{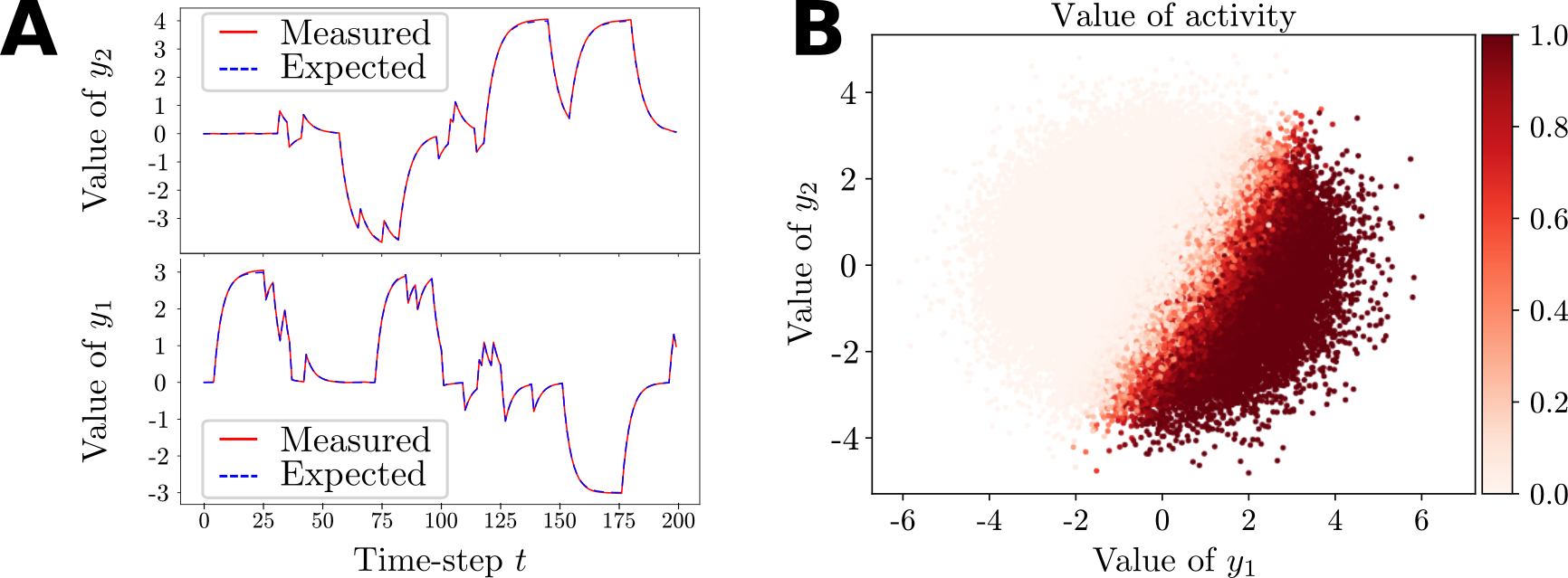}
    \caption{ Learning of $D$-dimensional integrators with sigmoidal networks. \textbf{A}: Comparison between expected and measured output on structured test sequences, designed to alternate between bursts of $\pm 1$ inputs and long periods with no external input to allow for visual discrimination of the origin of errors between scale and decay. \textbf{B}: Activity of a representative neuron in the $(y_1, y_2)$ plane, measured on white-noise inputs. The decays are equal to $0.8$ and $0.75$, $n=1000$, and the sigmoidal networks were trained using the proxy loss (\ref{eq:general_proxy}).}
    \label{fig:sigmoid_D2}
\end{figure}

\paragraph{Case of sigmoidal units.}
We have repeated the above analysis on networks with sigmoidal units, trained both from the batch and proxy losses. Results for a representative networks trained with the proxy loss to integrate $D=2$ channels are shown in Figure \ref{fig:sigmoid_D2}A. We observe an excellent match between the output integrals and their target values. Similar results, albeit less accurate are obtained with the batch loss.

As in the ReLU case, the connectivity matrix $\mW$ is characterized by $D$ large singular values, and a bulk of smaller ones. This bulk is influenced by several factors, including the initial condition over the matrix $W$ and the choice of the learning algorithm. Despite the presence of these small singular values, the $D$-dimensional nature of the current can be assessed, see ratios $r$ reported in Table \ref{fig:orth_norm_ratio}. The values of $r$ are much smaller than what would be expected from a null model, and confirm the low-dimensionality of the current manifold. Not suprisingly, the values of the ratios for sigmoidal networks are 2 to 10 times larger than for their ReLU counterparts (for the same size $n$), as expected from the higher difficulty to solve conditions (\ref{cond1},\ref{cond2}), see discussion in Section \ref{sec43}.

\paragraph{Nature of single neuron activity and mixed selectivity.}
The above findings allow us to determine how the state $h_i$ of a neuron depends on the integrals $\bar{\vy} = (\bar{y}_1, \bar{y}_2,..., \bar{y}_D)$ in a ReLU network:
\begin{equation}\label{rel3}
    h_i= \R{ \vs_i\trs \bar{\vy}}\ ,\ \text{with}\ \ s_{i,c}= \sum_{c'} R_{c',c} \,\vl_{c',i} \ .
\end{equation}

\begin{figure}
\centering
    \includegraphics[width=\textwidth]{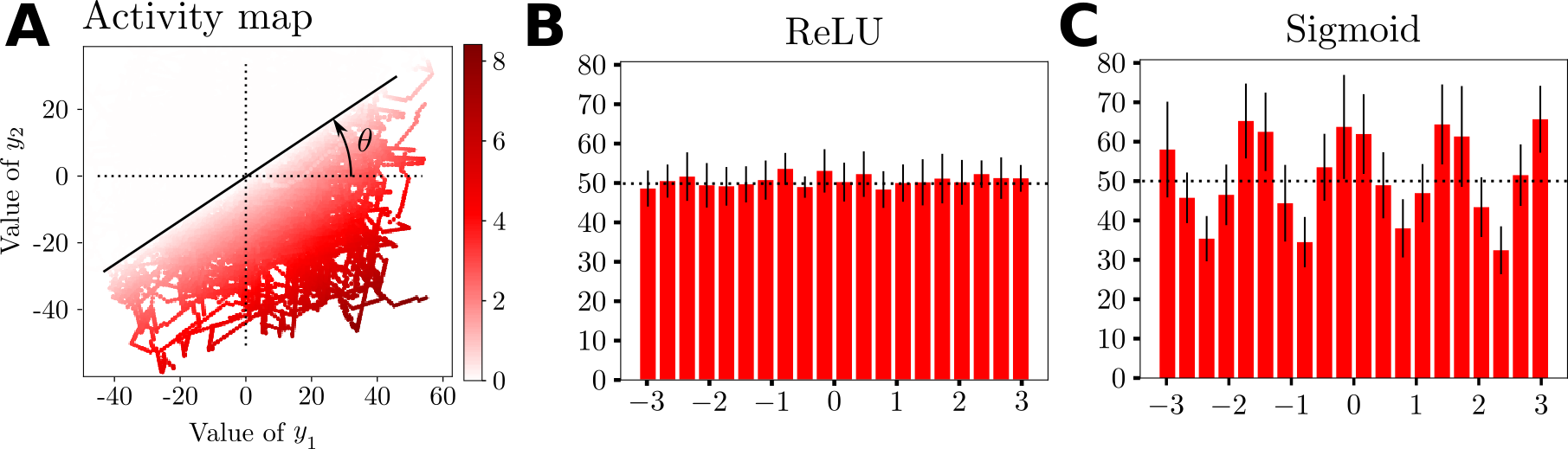}
    \caption{ Mixed selectivity in bichannel integrators. \textbf{A}: Activity $h_i$ of a representative neuron $i$ in a ReLU network as a function of the two integrals, aggregated across $512$ epochs of $T^{test}=200$ time-steps. This activity is of the form $\max(\vs_i\trs\vy, 0)$, meaning that the neuron will only ever be active in half of the $(y_1, y_2)$ plane. \textbf{B}: Distribution of the angle of the boundary plane between zero and non-zero activity across the $n=1000$ neurons of a ReLU network. \textbf{C}: Distribution of the angle in the case of a Sigmoidal network. Horizontal dotted lines represent the uniform distribution. Same parameters as in Figure \ref{fig:sigmoid_D2}; histograms were aggregated across 16 repetitions of the training.
    \label{fig:mixed_specificity}}
\end{figure}

From a geometrical point of view, as illustrated in Figure \ref{fig:mixed_specificity}A in the $D=2$ case, each neuron activity $h_i$ is the image through the ReLU non-linearity of the dot product between an associated direction $\vs_i$ and the set of  integrals ${\bar{\vy}}$. The same feature is encountered for sigmoidal units, as shown in Figure~\ref{fig:sigmoid_D2}B. We have then characterized the distribution of the angular direction of $\vs_i$ across the $n$ neurons, and find that it is equally distributed on $[0, 2\pi]$ when the network activation is ReLU, while it shows clear peaks for multiples of $\pi/2$ in the case of sigmoidal activation, see Figure \ref{fig:mixed_specificity}B\&C.
% This flat distribution in the case of ReLU neurons can be interpreted in light of the maximum entropy principle: each neuron carries information about a (non-linear) projection of the integral vector, and no specific orientation for this projection should be expected to be favored in the generic case where the encoders and decoders are unstructured. However, the choice of activation function can perturb this picture and encourage a non-uniform distribution of selectivity angles, meaning that this intuitive picture is incomplete.

In the solutions empirically obtained through Gradient Descent, either on the batch loss or the proxy loss and for any number $D$ of channels, we found that the network jointly encodes information about all integrals in the state of all neurons, a phenomenon similar to the one of "mixed selectivity" used to interpret cortical recordings in the field of computational neuroscience \citep{MixedSelectivity}, and closely related to the issue of class selectivity in computer vision, see \citep{leavitt_selectivity_2020}.

\begin{figure}[h!]
    \centering
    \includegraphics[width=\textwidth]{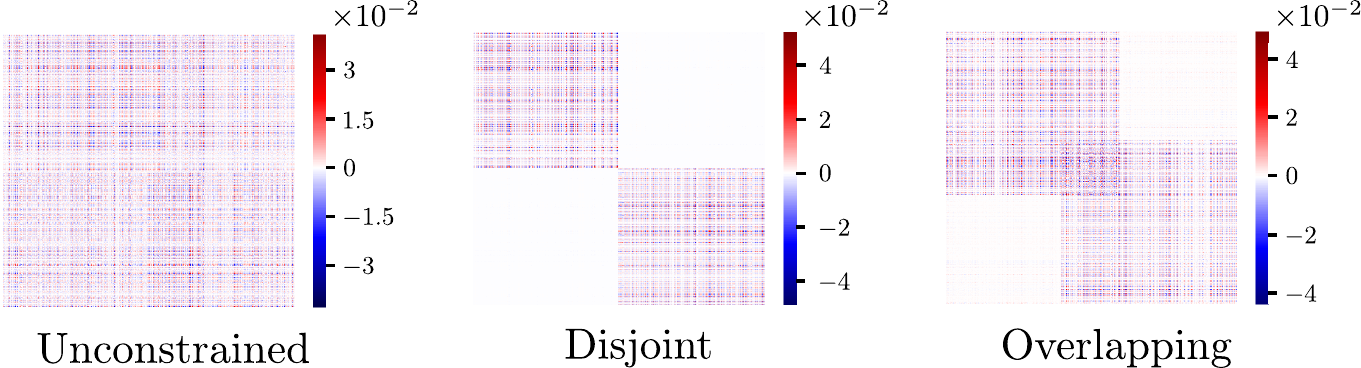}
        \caption{ Visualization of the elements of the weight matrix $\mW$ after training a ReLU network to integrate $D=2$ signals through batch--SGD in three different cases of initialization: (left) the encoders and decoders are independent Gaussian vectors without any restriction; (middle) the population is divided in two: the first half of the neurons have non-zero encoder and decoder only on channel 1, and similarly the other half on channel 2; (right) starting from the non-overlapping case, we allow a small fraction of the neurons (middle portion) to have non-zero components on all $\ve, \vd$ vectors. We find that the use of disjoint supports produces block-diagonal solutions where one population is in charge of one integral and isolated from the others, thus exhibiting single selectivity.}
        \label{fig:specificity_support}
\end{figure}

Mixed selectivity can be seen here as being deeply connected to the choice of the input and output layers of the network: in our experiments, all encoders and decoders have non-zero components on all neurons of the internal state. Therefore, during training, the connectivity matrix will be optimized in such a way that each of those neurons will extract and represent information about all integrals.
If we instead constrain the encoder and decoder for each channel to have the same support, spanning only $n/D$ neurons and non-overlapping with the support for any other channel, we find that the obtained solutions do not exhibit mixed selectivity anymore: the connection matrices $\mW$ are block-diagonal, indicating that the network subdivided into $D$ independent populations, each responsible for the coding of one integral. Relaxing the support constraint on either the encoders or the decoders causes mixed specificity to reappear. Last of all, allowing the support of the channels to overlap causes the corresponding neurons to exhibit mixed selectivity, while the rest of the network remains simply selective. Those findings are illustrated in Figure \ref{fig:specificity_support}.

We interpret this difference in behavior by the fact that the heavy constraints imposed between the encoders and decoders through their supports are enough to modify the energy landscape in such a way that the entropically favored connectivity matrices do not exhibit mixed selectivity anymore. None of these support constraints significantly impacts the final performance, nor the learning dynamics, and only the topology of the connectivity matrix is affected. Finally, we note that the choice of activation function also influences the distribution of selectivity angle, a fact that can not easily be understood from entropic considerations and could potentially be related to learning.

\paragraph{Learning with sign-constrained connections.}
So far, the only biological constraint we have considered regarded the states of neurons, which were forced to remain positive through the use of the ReLU activation function in order to represent firing rates. We now introduce a constraint on the weight matrix $\mW$ itself, corresponding to the observed division between excitatory and inhibitory neurons known as Dale's Law \citep{FundamentalNeuroscience}: at initialization, we fix a certain fraction of the columns of $\mW$, corresponding to the outgoing connections from a subpopulation of neurons, to have only negative entries, while the rest of the columns will have only positive entries. In order to maintain these constraints satisfied during training, after each step of optimization, we fix to $0$ all the elements of $\mW$ that changed sign.

\begin{figure}
\centering
    \includegraphics{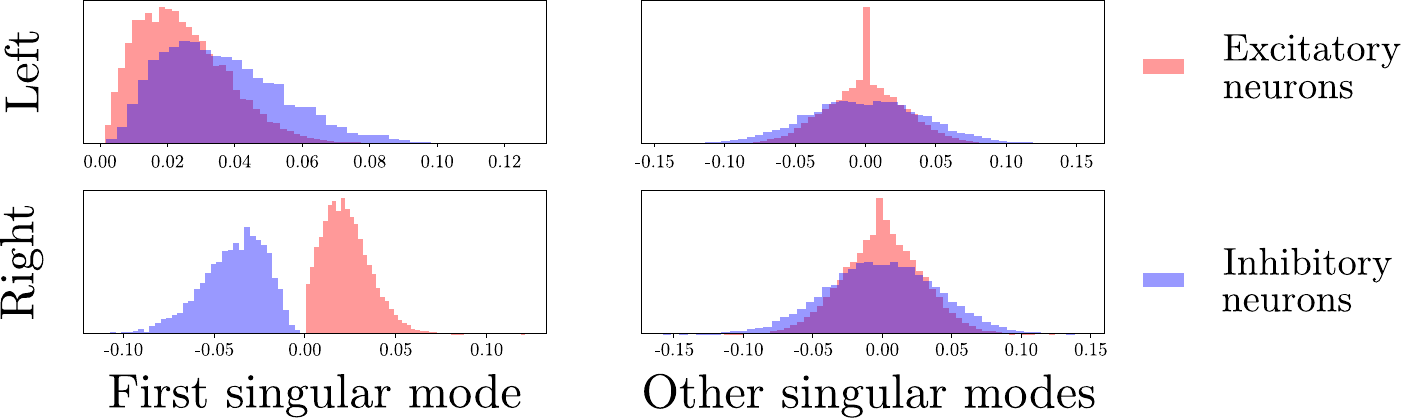}
    \caption{ Distribution of the components of the left and right singular vectors for the largest singular value (left) and the following $D$ ones (right). These histograms were obtained with 16 realizations of the batch--SGD training, using $n=1000$, $D=2$, and $25\%$ of inhibitory neurons. While the signs of the components of the $2$nd and $3$rd singular vectors appear random, they have a particular structure in the first singular vector : the left singular vector is always positive, while the right is positive (resp. negative) if the neuron is in the excitatory (resp. inhibitory) population; the corresponding rank-1 matrix has columns of fixed signs corresponding to the ones of Dale's constraints.}
    \label{fig:dale_modes}
\end{figure}

At the end of the training the weight matrices exhibit one additional relevant singular value compared to their unconstrained counterparts:
\begin{equation*}
    \mW \simeq  \sigma_{0}\, \vl_{0}\,\vr_{0}\trs + \sum_{c=1}^{D}\sigma_{c}\, \vl_{c}\,\vr_{c}\trs\ .
\end{equation*}
The rank-1 contribution coming from this additional mode has the correct signs to satisfy Dale's constraint, as illustrated in Figure \ref{fig:dale_modes}. Additionally, the left singular vector $\vl _0$ is almost orthogonal to all decoding vectors $\vd_c$, suggesting that this mode is not used for the computation of the integrals, but only as a way to satisfy the sign constraints over $\mW$. It should be noted that our empirical result does not rule out the existence of networks of rank $D$ performing $D$ multiplexed integrals while satisfying Dale's Law.  However, such solutions, if they exist, are not obtained through a simple Gradient Descent procedure from a zero or small $\mW$.

\section{Conclusion and perspectives}
\addcontentsline{toc}{section}{Conclusion}
\label{sec:conclusion}

\paragraph{Summary of results and open questions.}
We have studied in this work how a RNN with $n$ neurons learns to perform one or more integrations of temporal inputs; each integration was characterized by the target values of the scale factor $s$ and of the decay coefficient $\gamma$ (generally, slightly below 1).

In the case of a RNN with linear activation performing a single integral, we have precisely characterized the length of the temporal input necessary for perfect generalization (integration of any temporal signal), the optimal learning rate and the convergence time of the training procedure when the weight matrix is initially set to zero (or is small enough in norm). The coding of the integral is realized in a simple way: the activity vector of the entire neural population varies along a 1-dimensional direction in the $n$-dimensional space, with a proportionality factor equal to the integral.

In the case of ReLU activation, very accurate integration was obtained at the end of the training too. While a full mathematical analysis seemed much harder than for linear activation, we showed empirical evidence for the fact that the activity vector belongs to a piecewise 1-dimensional manifold. Coding of the positive and negative values of the integrals is done by two essentially non-overlapping populations of neurons, switching on and off when the integral value crosses zero. Remarkably, the pre-activation current of the ReLU units shows a simple behaviour: it is proportional to the integral. We have derived sufficient conditions over the weight matrix for such a coding to take place, and characterized the nature (directions of left and right eigenvectors, amplitude of singular value as a function of $s, \gamma$) of the corresponding rank-1 integrator.

In the case of a multiplexed network with  $D$ input/output channels, we have found that the weight matrix is of rank $D$; this statement is exact for linear activation and approximately true for ReLU activation RNNs, whose weight matrix has $D$ large singular values compared to the $n-D$ remaining ones. Consequently, the network activity is restricted to a $D$--dimensional manifold in $\mathbb{R}^n$, whose geometry is imposed by the activation function of the neurons. For ReLU activation, as in the single-integral case, strong empirical evidence suggests that the pre-activation currents are linear combinations of the $D$ integrals and span a $D$-dimensional linear subspace.

It is important to stress that the above results are not mere consequences of the threshold-linear nature of ReLU units. We have repeated our analysis with saturating units, obeying a sigmoidal activation function, with essentially the same results. Interestingly, some units never saturate for all possible values of the integral(s), other do, and all participate to produce the right outputs. To elucidate the reason for the $D$-dimensional nature of the coding of integrals by the currents, we have introduced a proxy loss reflecting sufficient conditions for such a coding. The networks trained from data (and the batch loss) behave similarly to the networks minimizing this proxy loss, both from the point of view of performance and representations.

From a purely machine-learning point of view, our work shows the versality of RNNs to achieve simultaneously several computational tasks. The variety of representations supporting these computations could then be harnessed for transfer learning, see \citep{TransferLearningPan} for a review, by using our trained RNN as a (possibly fixed) feature extractor. One example of such a task is the one of context-dependent integration, studied in the prefrontal cortex of monkeys by \citep{mante_context-dependent_2013}, and which we adapt to our setup in Appendix \ref{app:mante}. The proxy loss we derived could also \textit{a priori} be used as part of a full-task loss, following a similar reasoning to \citep{haviv_understanding_2019}, where one term in the loss is used to encourage internal dynamics that are known to be relevant for the task at hand and facilitate training.

Empirical analysis shows that very accurate multi-integrators with non-trivial activation functions can be obtained through Gradient Descent, and the representation scheme they adapt is linear in the space of currents. Three main limitations in this observation have to be noted. First, we do not show that this is the only representation scheme possible, and different solutions could possibly be found from pure mathematical reasoning. Second, rigorous analysis of the proxy loss remains necessary to understand in which conditions these representations are achievable, and to which accuracy. Finally, our study has focused on the case the case where the number of integrals $D$ is small and the number of neurons $n$ is large, and the question of how the optimal computational capacity (maximal sustainable value of $D$) precisely increases with $n$ remains to be understood in the case of RNNs with non-linear activation.

%Similarly, it would be interesting to better characterize the presence of the small singular values in the spectrum of $\mW$, and understand whether or not they are an artifact of the large but finite $n$ value.

% Last of all, while we have empirically shown that GD can generate very accurate multi-integrators in , how the optimal computational capacity (maximal sustainable value of $D$) precisely increases with $n$ remains to be understood in the case of RNNs with non-linear activation.

\paragraph{Nature of representations and connection with computational neuroscience.}
While scalar integration using a single-layer recurrent network is far from state-of-the-art Machine Learning, the abundance of studies in the field of neuroscience (often motivated by the oculo-motor system in fish) and the absence of a comprehensive theory of representation in such networks make it a worthwhile case study. Our theoretical analysis provide new evidence for the relevance of low-dimensional representations, and this result is robust to changes in the training method, the initial conditions of the weight matrix, as well as the choice of activation function. Our work therefore provides additional motivation for the theoretical study of the properties of RNNs with low-rank coupling matrix initiated in the contexts of statistical physics \citep{LowRankRNNs} and computational neuroscience, see \citep{barak_recurrent_2017} for a review.

As far as neuroscience is concerned, we believe that our result about the encoding of multiple integrals by each neuron, expressed by (\ref{rel3}), is of particular interest. There is, indeed, a very striking analogy between our findings and the concept of mixed selectivity used to interpret cortical recordings in the field of computational neuroscience \citep{MixedSelectivity}. For a long time, neuroscientists have focused on neurons whose activities  depended on a single sensory relevant variable, such as the orientation of a bar in the visual cortex area V1 or the animal's head direction in the subiculum (in our case, the value of one particular integral $y_c$). Such neurons are, obviously,  easier to identify from activity recordings. However, there is growing recognition that most cells display mixed sensitivity, that is, have activities varying non-linearly with several relevant variables, and that the relative degree of importance of each variable in determining the activity may considerably vary from neuron to neuron  (as we find in Figure \ref{fig:mixed_specificity}). Such mixed representations could be useful for decision making based on multi-sensorial streams of information, a possibility sometimes put forward to explain their relevance in neuroscience. It is, from this point of view, remarkable that mixed representations spontaneously emerge in our study, where the RNN lacks any explicit incentive to exploit them, simply because they are much more likely than pure representations when the encoders and decoders have no intrinsic structure (Figure \ref{fig:mixed_specificity}). The computational advantages of such mixed representations have been studied in \citep{leavitt_selectivity_2020, leavitt_relationship_2020}, and suggest that they could improve both generalization and robustness of the performed computations. Other studies have focused on the emergence of disentangled representations, which have been shown to be relevant in both Natural Language Processing \citep{radford_learning_2017} and Computer Vision \citep{denton_unsupervised_2017, lee_diverse_2018}, suggesting that the optimal type of representation might depend on the specific task it supports. Studying the representations of computational tasks in artificial neural networks could therefore be a valuable tool to understand their biological counterparts, an approach already proposed in the domain of spatial navigation \citep{Banino}.

\vskip .3cm
\noindent {\bf Acknowledgements.} We are grateful to the referees for useful suggestions, in particular the connection with \cite{mante_context-dependent_2013}, see Appendix \ref{app:mante}, as well as the study of selectivity with sigmoidal networks, which shows preferred orientations, see Figure \ref{fig:mixed_specificity}C. We benefited from the support of NVIDIA Corporation with the donation of a Tesla K40 GPU card.

%Maybe with more layers (belief type?) we can generate more complex selectivity maps

\newpage
% \bibliographystyle{style/apa}
% \bibliography{bibliography.bib}

\begin{thebibliography}{}

\bibitem[\protect\astroncite{Aksay et~al.}{2007}]{GoldfishVPNI}
Aksay, E., Olasagasti, I., Mensh, B.~D., Baker, R., Goldman, M.~S., and Tank,
  D.~W. (2007).
\newblock Functional dissection of circuitry in a neural integrator.
\newblock {\em Nature Neuroscience}, 10(4):494--504.

\bibitem[\protect\astroncite{Almagro~Armenteros et~al.}{2017}]{DeepLoc}
Almagro~Armenteros, J.~J., Sonderby, C.~K., Sonderby, S.~K., Nielsen, H., and
  Winther, O. (2017).
\newblock {DeepLoc}: prediction of protein subcellular localization using deep
  learning.
\newblock {\em Bioinformatics}, 33(21):3387--3395.

\bibitem[\protect\astroncite{Amodei et~al.}{2015}]{DeepSpeech}
Amodei, D., Anubhai, R., Battenberg, E., Case, C., Casper, J., Catanzaro, B.,
  Chen, J., Chrzanowski, M., Coates, A., Diamos, G., Elsen, E., Engel, J., Fan,
  L., Fougner, C., Han, T., Hannun, A., Jun, B., LeGresley, P., Lin, L.,
  Narang, S., Ng, A., Ozair, S., Prenger, R., Raiman, J., Satheesh, S.,
  Seetapun, D., Sengupta, S., Wang, Y., Wang, Z., Wang, C., Xiao, B., Yogatama,
  D., Zhan, J., and Zhu, Z. (2015).
\newblock Deep {Speech} 2: {End}-to-{End} {Speech} {Recognition} in {English}
  and {Mandarin}.
\newblock {\em arXiv:1512.02595 [cs]}.
\newblock arXiv: 1512.02595.

\bibitem[\protect\astroncite{Arora et~al.}{2019}]{ImplicitRegularization}
Arora, S., Cohen, N., Hu, W., and Luo, Y. (2019).
\newblock Implicit regularization in deep matrix factorization.
\newblock {\em CoRR}, abs/1905.13655.

\bibitem[\protect\astroncite{Banino et~al.}{2018}]{Banino}
Banino, A., Barry, C., Uria, B., Blundell, C., Lillicrap, T., Mirowski, P.,
  Pritzel, A., Chadwick, M.~J., Degris, T., Modayil, J., Wayne, G., Soyer, H.,
  Viola, F., Zhang, B., Goroshin, R., Rabinowitz, N., Pascanu, R., Beattie, C.,
  Petersen, S., Sadik, A., Gaffney, S., King, H., Kavukcuoglu, K., Hassabis,
  D., Hadsell, R., and Kumaran, D. (2018).
\newblock Vector-based navigation using grid-like representations in artificial
  agents.
\newblock {\em Nature}, 557(7705):429.

\bibitem[\protect\astroncite{Barak}{2017}]{barak_recurrent_2017}
Barak, O. (2017).
\newblock Recurrent neural networks as versatile tools of neuroscience
  research.
\newblock {\em Current Opinion in Neurobiology}, 46:1--6.

\bibitem[\protect\astroncite{Chung et~al.}{2014}]{EmpiricalRNN}
Chung, J., Gulcehre, C., Cho, K., and Bengio, Y. (2014).
\newblock Empirical {Evaluation} of {Gated} {Recurrent} {Neural} {Networks} on
  {Sequence} {Modeling}.
\newblock {\em arXiv:1412.3555 [cs]}.
\newblock arXiv: 1412.3555.

\bibitem[\protect\astroncite{Collins et~al.}{2017}]{collins_capacity_2017}
Collins, J., Sohl-Dickstein, J., and Sussillo, D. (2017).
\newblock Capacity and {Trainability} in {Recurrent} {Neural} {Networks}.
\newblock {\em arXiv:1611.09913 [cs, stat]}.
\newblock arXiv: 1611.09913.

\bibitem[\protect\astroncite{Denton and
  Birodkar}{2017}]{denton_unsupervised_2017}
Denton, E.~L. and Birodkar, v. (2017).
\newblock Unsupervised {Learning} of {Disentangled} {Representations} from
  {Video}.
\newblock In Guyon, I., Luxburg, U.~V., Bengio, S., Wallach, H., Fergus, R.,
  Vishwanathan, S., and Garnett, R., editors, {\em Advances in {Neural}
  {Information} {Processing} {Systems} 30}, pages 4414--4423. Curran
  Associates, Inc.

\bibitem[\protect\astroncite{Elman}{1990}]{Elman_RNN}
Elman, J.~L. (1990).
\newblock Finding structure in time.
\newblock {\em Cognitive Science}, 14(2):179--211.

\bibitem[\protect\astroncite{Haviv et~al.}{2019}]{haviv_understanding_2019}
Haviv, D., Rivkind, A., and Barak, O. (2019).
\newblock Understanding and {Controlling} {Memory} in {Recurrent} {Neural}
  {Networks}.
\newblock In {\em International {Conference} on {Machine} {Learning}}, pages
  2663--2671. PMLR.

\bibitem[\protect\astroncite{Kingma and Ba}{2017}]{Adam}
Kingma, D.~P. and Ba, J. (2017).
\newblock Adam: {{A Method}} for {{Stochastic Optimization}}.
\newblock {\em arXiv:1412.6980 [cs]}.

\bibitem[\protect\astroncite{Leavitt and
  Morcos}{2020a}]{leavitt_selectivity_2020}
Leavitt, M.~L. and Morcos, A. (2020a).
\newblock Selectivity considered harmful: evaluating the causal impact of class
  selectivity in {DNNs}.
\newblock {\em arXiv:2003.01262 [cs, q-bio, stat]}.
\newblock arXiv: 2003.01262.

\bibitem[\protect\astroncite{Leavitt and
  Morcos}{2020b}]{leavitt_relationship_2020}
Leavitt, M.~L. and Morcos, A.~S. (2020b).
\newblock On the relationship between class selectivity, dimensionality, and
  robustness.
\newblock {\em arXiv:2007.04440 [cs, stat]}.
\newblock arXiv: 2007.04440.

\bibitem[\protect\astroncite{Lee et~al.}{1997}]{SeungNonLinear}
Lee, D.~D., Reis, B.~Y., Seung, H.~S., and Tank, D.~W. (1997).
\newblock Nonlinear {{Network Models}} of the {{Oculomotor Integrator}}.
\newblock In Bower, J.~M., editor, {\em Computational {{Neuroscience}}}, pages
  371--377. {Springer US}, {Boston, MA}.

\bibitem[\protect\astroncite{Lee et~al.}{2018}]{lee_diverse_2018}
Lee, H.-Y., Tseng, H.-Y., Huang, J.-B., Singh, M., and Yang, M.-H. (2018).
\newblock Diverse {Image}-to-{Image} {Translation} via {Disentangled}
  {Representations}.
\newblock pages 35--51.

\bibitem[\protect\astroncite{Li et~al.}{2017}]{li_understanding_2017}
Li, J., Monroe, W., and Jurafsky, D. (2017).
\newblock Understanding {Neural} {Networks} through {Representation} {Erasure}.
\newblock {\em arXiv:1612.08220 [cs]}.
\newblock arXiv: 1612.08220.

\bibitem[\protect\astroncite{Lipton et~al.}{2015}]{RNNReview}
Lipton, Z.~C., Berkowitz, J., and Elkan, C. (2015).
\newblock A {{Critical Review}} of {{Recurrent Neural Networks}} for {{Sequence
  Learning}}.
\newblock {\em arXiv:1506.00019 [cs]}.

\bibitem[\protect\astroncite{Luong et~al.}{2016}]{MultitaskSequenceSequence}
Luong, M.-T., Le, Q.~V., Sutskever, I., Vinyals, O., and Kaiser, L. (2016).
\newblock Multi-task {{Sequence}} to {{Sequence Learning}}.
\newblock {\em arXiv:1511.06114 [cs, stat]}.

\bibitem[\protect\astroncite{Mante et~al.}{2013}]{mante_context-dependent_2013}
Mante, V., Sussillo, D., Shenoy, K.~V., and Newsome, W.~T. (2013).
\newblock Context-dependent computation by recurrent dynamics in prefrontal
  cortex.
\newblock {\em Nature}, 503(7474):78--84.

\bibitem[\protect\astroncite{Mastrogiuseppe and Ostojic}{2018}]{LowRankRNNs}
Mastrogiuseppe, F. and Ostojic, S. (2018).
\newblock Linking connectivity, dynamics and computations in low-rank recurrent
  neural networks.
\newblock {\em Neuron}, 99(3):609--623.e29.

\bibitem[\protect\astroncite{Montavon et~al.}{2018}]{InterpretingMontavon}
Montavon, G., Samek, W., and M{\"u}ller, K.-R. (2018).
\newblock Methods for interpreting and understanding deep neural networks.
\newblock {\em Digital Signal Processing}, 73:1--15.

\bibitem[\protect\astroncite{Olah et~al.}{2018}]{InterpretingInteractive}
Olah, C., Satyanarayan, A., Johnson, I., Carter, S., Schubert, L., Ye, K., and
  Mordvintsev, A. (2018).
\newblock The {{Building Blocks}} of {{Interpretability}}.
\newblock {\em Distill}, 3(3):e10.

\bibitem[\protect\astroncite{Pan and Yang}{2010}]{TransferLearningPan}
Pan, S.~J. and Yang, Q. (2010).
\newblock A {{Survey}} on {{Transfer Learning}}.
\newblock {\em IEEE Transactions on Knowledge and Data Engineering},
  22(10):1345--1359.

\bibitem[\protect\astroncite{Paszke et~al.}{2019}]{pytorch}
Paszke, A., Gross, S., Massa, F., Lerer, A., Bradbury, J., Chanan, G., Killeen,
  T., Lin, Z., Gimelshein, N., Antiga, L., Desmaison, A., Kopf, A., Yang, E.,
  DeVito, Z., Raison, M., Tejani, A., Chilamkurthy, S., Steiner, B., Fang, L.,
  Bai, J., and Chintala, S. (2019).
\newblock Pytorch: An imperative style, high-performance deep learning library.
\newblock In Wallach, H., Larochelle, H., Beygelzimer, A., d\textquotesingle
  Alch\'{e}-Buc, F., Fox, E., and Garnett, R., editors, {\em Advances in Neural
  Information Processing Systems 32}, pages 8024--8035. Curran Associates, Inc.

\bibitem[\protect\astroncite{Radford et~al.}{2017}]{radford_learning_2017}
Radford, A., Jozefowicz, R., and Sutskever, I. (2017).
\newblock Learning to {Generate} {Reviews} and {Discovering} {Sentiment}.
\newblock {\em arXiv:1704.01444 [cs]}.
\newblock arXiv: 1704.01444.

\bibitem[\protect\astroncite{Richards et~al.}{2019}]{richards_deep_2019}
Richards, B.~A., Lillicrap, T.~P., Beaudoin, P., Bengio, Y., Bogacz, R.,
  Christensen, A., Clopath, C., Costa, R.~P., de~Berker, A., Ganguli, S.,
  Gillon, C.~J., Hafner, D., Kepecs, A., Kriegeskorte, N., Latham, P., Lindsay,
  G.~W., Miller, K.~D., Naud, R., Pack, C.~C., Poirazi, P., Roelfsema, P.,
  Sacramento, J., Saxe, A., Scellier, B., Schapiro, A.~C., Senn, W., Wayne, G.,
  Yamins, D., Zenke, F., Zylberberg, J., Therien, D., and Kording, K.~P.
  (2019).
\newblock A deep learning framework for neuroscience.
\newblock {\em Nature Neuroscience}, 22(11):1761--1770.

\bibitem[\protect\astroncite{Rigotti et~al.}{2013}]{MixedSelectivity}
Rigotti, M., Barak, O., Warden, M.~R., Wang, X.-J., Daw, N.~D., Miller, E.~K.,
  and Fusi, S. (2013).
\newblock The importance of mixed selectivity in complex cognitive tasks.
\newblock {\em Nature}, 497(7451):585--590.

\bibitem[\protect\astroncite{Robinson}{1989}]{NeuralIntegratorRobinson}
Robinson, D.~A. (1989).
\newblock Integrating with {{Neurons}}.
\newblock {\em Annual Review of Neuroscience}, 12(1):33--45.

\bibitem[\protect\astroncite{Saxe et~al.}{2014}]{saxe_exact_2014}
Saxe, A.~M., McClelland, J.~L., and Ganguli, S. (2014).
\newblock Exact solutions to the nonlinear dynamics of learning in deep linear
  neural networks.
\newblock {\em arXiv:1312.6120 [cond-mat, q-bio, stat]}.
\newblock arXiv: 1312.6120.

\bibitem[\protect\astroncite{Schuessler
  et~al.}{2020a}]{schuessler_dynamics_2020}
Schuessler, F., Dubreuil, A., Mastrogiuseppe, F., Ostojic, S., and Barak, O.
  (2020a).
\newblock Dynamics of random recurrent networks with correlated low-rank
  structure.
\newblock {\em Physical Review Research}, 2(1):013111.
\newblock arXiv: 1909.04358.

\bibitem[\protect\astroncite{Schuessler
  et~al.}{2020b}]{schuessler_interplay_2020}
Schuessler, F., Mastrogiuseppe, F., Dubreuil, A., Ostojic, S., and Barak, O.
  (2020b).
\newblock The interplay between randomness and structure during learning in
  {RNNs}.
\newblock {\em arXiv:2006.11036 [q-bio]}.
\newblock arXiv: 2006.11036.

\bibitem[\protect\astroncite{Seung}{1996}]{SeungLinear}
Seung, H.~S. (1996).
\newblock How the brain keeps the eyes still.
\newblock {\em Proceedings of the National Academy of Sciences},
  93(23):13339--13344.

\bibitem[\protect\astroncite{Song et~al.}{2016}]{SongWangEINets}
Song, H.~F., Yang, G.~R., and Wang, X.-J. (2016).
\newblock Training excitatory-inhibitory recurrent neural networks for
  cognitive tasks: {{A}} simple and flexible framework.
\newblock {\em PLOS Computational Biology}, 12(2):1--30.

\bibitem[\protect\astroncite{Squire et~al.}{2012}]{FundamentalNeuroscience}
Squire, L., Berg, D., Bloom, F., Lac, S., Ghosh, A., and Spitzer, N. (2012).
\newblock {\em Fundamental {{Neuroscience}}: {{Fourth Edition}}}.

\bibitem[\protect\astroncite{Sussillo and Barak}{2012}]{sussillo_opening_2012}
Sussillo, D. and Barak, O. (2012).
\newblock Opening the {Black} {Box}: {Low}-{Dimensional} {Dynamics} in
  {High}-{Dimensional} {Recurrent} {Neural} {Networks}.
\newblock {\em Neural Computation}, 25(3):626--649.

\bibitem[\protect\astroncite{Tanaka et~al.}{2019}]{ReservoirReview}
Tanaka, G., Yamane, T., H{\'e}roux, J.~B., Nakane, R., Kanazawa, N., Takeda,
  S., Numata, H., Nakano, D., and Hirose, A. (2019).
\newblock Recent advances in physical reservoir computing: {{A}} review.
\newblock {\em Neural Networks}, 115:100--123.

\bibitem[\protect\astroncite{Virtanen et~al.}{2020}]{scipy}
Virtanen, P., Gommers, R., Oliphant, T.~E., Haberland, M., Reddy, T.,
  Cournapeau, D., Burovski, E., Peterson, P., Weckesser, W., Bright, J., {van
  der Walt}, S.~J., Brett, M., Wilson, J., Millman, K.~J., Mayorov, N., Nelson,
  A. R.~J., Jones, E., Kern, R., Larson, E., Carey, C.~J., Polat, {\.I}., Feng,
  Y., Moore, E.~W., VanderPlas, J., Laxalde, D., Perktold, J., Cimrman, R.,
  Henriksen, I., Quintero, E.~A., Harris, C.~R., Archibald, A.~M., Ribeiro,
  A.~H., Pedregosa, F., and {van Mulbregt}, P. (2020).
\newblock {{SciPy}} 1.0: Fundamental algorithms for scientific computing in
  {{Python}}.
\newblock {\em Nature Methods}, 17(3):261--272.

\bibitem[\protect\astroncite{Wong and Wang}{2006}]{WangDecisionMaking}
Wong, K.-F. and Wang, X.-J. (2006).
\newblock A {{Recurrent Network Mechanism}} of {{Time Integration}} in
  {{Perceptual Decisions}}.
\newblock {\em Journal of Neuroscience}, 26(4):1314--1328.

\bibitem[\protect\astroncite{Zhang and Zhu}{2018}]{InterpretingZhang}
Zhang, Q.-s. and Zhu, S.-c. (2018).
\newblock Visual interpretability for deep learning: A survey.
\newblock {\em Frontiers of Information Technology \& Electronic Engineering},
  19(1):27--39.

\end{thebibliography}

\appendix
\newpage
\setcounter{section}{0}
% {\LARGE \begin{center} Low-dimensional manifolds support multiplexed\\ integrations in~Recurrent~Neural~Networks \end{center}\\
{\LARGE\bf Supplementary information}

% \ \\
% {\bf \large Arnaud Fanthomme$^{\displaystyle 1}$, Rémi Monasson$^{\displaystyle 1}$}\\
% {$^{\displaystyle 1}$Laboratoire de Physique de l'Ecole Normale Superieure, Paris, France}\\

% \maketitle

\section{Fully averaged loss for linear single-channel integrators}
    \label{app:linear_loss}
    For an arbitrary input sequence $(x_t)_{0\leq t\leq T-1}$, we compute through induction the value of the output at any time $t$ as:
    \begin{equation*}
        \forall t \in \mathbb{N}, \, y_t = \sum_{q=0}^t x_{t-q} \vd^T\mW^{q+1}\ve := \sum_{q=0}^t x_{t-q} \mu_{q+1}
    \end{equation*}

   \flushleft The target output is $\overline{y}_t = s \sum_{q=0}^t x_{t-q} \gamma^{q+1}$, so that the square error is:
    \begin{equation*}
        \begin{split}
            \epsilon_t^2 &= (y_t-\overline{y}_t)^2 \\
            &= [\sum_{q=0}^t x_{t-q} (\mu_{q+1}-s\gamma^{q+1})]^2 \\
            &= \sum_{q, p=0}^t x_{t-q}x_{t-p} (\mu_{q+1}-s\gamma^{q+1})(\mu_{p+1}-s\gamma^{p+1}) \\
        \end{split}
    \end{equation*}
    The loss to minimize is the average of the sum of those errors along input sequences of length~$T$:
    \begin{equation}
        \begin{split}
            \mathcal{L}(\mW) &= \left \langle \sum_{t=0}^{T-1} \epsilon_t^2 \right \rangle = \left \langle \sum_{t=0}^{T-1} \sum_{p,q=0}^t x_{t-q}x_{t-p} (\mu_{q+1}-s\gamma^{q+1})(\mu_{p+1}-s\gamma^{p+1}) \right \rangle \\
            &=  \left \langle \sum_{t=0}^{T-1}  \sum_{p,q=0}^{T-1} x_{t-q}x_{t-p}\bm{1}_{q\leq t}\bm{1}_{p \leq t} \right \rangle (\mu_{q+1}-s\gamma^{q+1})(\mu_{p+1}-s\gamma^{p+1}) \\
            &= \sum_{p,q=0}^{T-1}  \left \langle \sum_{t=0}^{T-1} x_{t-q}x_{t-p} \bm{1}_{q\leq t}\bm{1}_{p \leq t} \right \rangle (\mu_{q+1}-s\gamma^{q+1})(\mu_{p+1}-s\gamma^{p+1}) \\
            &:=  \sum_{p,q=1}^{T} \bm{\chi}_{qp} (\mu_{q}-s\gamma^{q})(\mu_{p}-s\gamma^{p}) \\
        \end{split}
    \end{equation}
    where we introduced the time-integrated correlation matrix $\bm{\chi}$.

    $\bm{\chi}$ is symmetric, and it is easily shown that:
    \begin{equation*}
    \begin{split}
        \forall \vv \in \mathbb{R}^T, \vv\trs \bm{\chi} \vv &= \sum_{p,q=0}^{T-1} v_q \chi_{qp} v_p = \sum_{p,q=0}^{T-1} \left \langle \sum_{t=0}^{T-1} v_q x_{t-q}x_{t-p} \bm{1}_{q\leq t}\bm{1}_{p \leq t} v_p \right \rangle \\
        &= \left \langle \sum_{t=0}^{T-1} (\sum_{q=0}^{T-1}v_q x_{t-q} \bm{1}_{q\leq t})(\sum_{p=0}^{T-1}x_{t-p} \bm{1}_{p \leq t} v_p) \right \rangle \\
        &= \sum_{t=0}^{T-1}  \left\langle  \left(\sum_{p=0}^{t}x_{t-p} \, v_p \right)^2 \right\rangle \ .
    \end{split}
    \end{equation*}

    Therefore, $\bm{\chi}$ is non-negative. Assuming now that $\vv$ is such that the quadratic form above vanishes, the term corresponding to $t=0$, equal to $v_0^2\langle x_0^2 \rangle$ vanishes, entailing that $v_0=0$ as soon as the input is assumed to have positive probability to be non-zero at this first time-step.
    % (we assume that $t=0$ is the smallest time at which the input $x_t$ received by the network has a positive probability to be non null).
    Then, the $t=1$ contribution, $\langle( x_0 \,v_1 + x_1\, v_0)^2\rangle= \langle( x_0 \,v_1 )^2\rangle$  also vanishes, which implies that $v_1=0$. By recursion over $t$, all the components of $\vv$ must vanish, which shows that $\bm{\chi}$ is definite positive.

\section{Gradient and Hessian of the linear single channel loss}
\label{app:gradient_hessian_linear}
We have:
\begin{equation}
\begin{split}
     \nabla_{W_{ij}}\mathcal{L} &= \sum_{q,p=1}^{T} \chi_{qp} \left [(\mu_q-s\gamma^{q}) \frac{\partial \mu_p}{\partial W_{ij}} + (\mu_p-s\gamma^{p}) \frac{\partial \mu_q}{\partial W_{ij}} \right ]\\
     &= 2\sum_{q,p=1}^{T} \chi_{qp} (\mu_q-s\gamma^{q}) \frac{\partial \mu_p}{\partial W_{ij}}.
 \end{split}
\end{equation}

We compute through induction:
\begin{equation*}
     \frac{\partial W^p_{ij}}{\partial W_{kl}} = \sum_{m=0}^{p-1}W^{m}_{ik}W^{p-1-m}_{lj} \quad \text{hence} \quad \frac{\partial \mu_p}{\partial W_{kl}} = \sum_{i, j=1}^n\sum_{m=0}^{p-1}d_i W^{m}_{ik}W^{p-1-m}_{lj}e_j
\end{equation*}

So that the gradient of $\mathcal{L}$ with respect to $\mW$ is:
\begin{equation}
     \nabla_{\emW_{ij}}\mathcal{L} = 2 \sum_{q,p=1}^{T} \chi_{qp} (\mu_q-s\gamma^{q}) \sum_{m=0}^{p-1}\sum_{\alpha, \beta} (d_{\alpha}W^m_{\alpha i})(W^{p-1-m}_{j \beta}e_{\beta})
\end{equation}

We now want to compute the Hessian $\mathcal{H}$ of this loss:
\begin{equation*}
\begin{split}
    \mathcal{H}_{ij,kl} = \frac{\partial \mathcal{L}}{\partial W_{ij} \partial W_{kl}} &= 2 \sum_{q,p=1}^{T} \chi_{qp} (\mu_q-s\gamma^{q}) \frac{\partial }{\partial W_{kl}}\left [  \sum_{m=0}^{p-1}\sum_{\alpha, \beta} (d_{\alpha}W^m_{\alpha i})(W^{p-1-m}_{j \beta}e_{\beta}) \right ]\\
    &+ 2 \sum_{q,p=1}^{T} \chi_{qp} \frac{\partial \mu_q}{\partial W_{kl}}  \sum_{m=0}^{p-1}\sum_{\alpha, \beta} (d_{\alpha}W^m_{\alpha i})(W^{p-1-m}_{j \beta}e_{\beta})
\end{split}
\end{equation*}

We will only be interested in the value of the Hessian at global minima of $\mathcal{L}$, so that the first term in this equation will not contribute. In that case, we find:
\begin{equation}\label{eq:linear_hessian}
     \mathcal{H}_{ij,kl} = 2 \sum_{q,p=1}^{T} \chi_{qp} \left [\sum_{m=0}^{p-1}\sum_{\alpha, \beta} (d_{\alpha}W^m_{\alpha i})(W^{p-1-m}_{j \beta}e_{\beta})\right ] \left [ \sum_{\tilde{\a}, \tilde{\b}=1}^n\sum_{\tilde{m}=0}^{q-1}d_{\tilde{\a}} W^{\tilde{m}}_{\tilde{\a}k}W^{q-1-\tilde{m}}_{l\tilde{\b}}e_{\tilde{\b}} \right ]
\end{equation}

% \section{Global minima and Generalizing Integrators }
% \label{app:linear_global_minima}

\section{Two special cases of null initialization}

\label{app:special_cases}

\paragraph{Exact solution for $T=1$.}
    We begin by the simplest case possible, when the epochs are of length $1$. The gradient descent updates become in that case:
    \begin{equation*}
            \Delta W_{ij} = - 2 \eta (\vd\trs\mW \ve-s\gamma)d_i e_j
    \end{equation*}

    Hence, we have at any time $\mW= \w \vd \ve^T$, so that we can study the optimization dynamics on the scalar $\w$ only~:
    \begin{equation*}
            \Delta  \w = - 2 \eta (\w||\ve||^2||\vd||^2-s\gamma)
    \end{equation*}

    Therefore, after $\tau$ steps of optimization, the coefficient $\w$ is equal to \begin{equation*}
        \w^{(\tau)}=\frac{s \gamma}{||\ve||^2||\vd||^2} [1 - (1-2 \eta ||\ve||^2||\vd||^2)^{\tau}]
    \end{equation*}

    This dynamics is stable if and only if $\eta < ||\ve||^{-2}||\vd||^{-2}$. When it is, the network converges exponentially fast to $\mW = \frac{\gamma s}{||\ve||^2||\vd||^2}\vd \ve^T$, which gives the following moments~:

    \begin{equation*}
        \forall k\geq 1, \mu_k = \gamma s (\frac{\gamma s \, \ve\cdot\vd}{||\ve||^{2}||\vd||^{2}})^{k-1}
    \end{equation*}

    Therefore, in that case, we converge towards a solution that achieves the desired scaling $s$, but has a decay constant that is fixed by the initial choice of $s$, $\ve$ and $\vd$.

    \paragraph{Same encoder and decoder, $T>1$, uncorrelated inputs.}

    For this part, we assume that $\ve = \vd$. Considering that the first update in that case is proportional to $\ve\ve^T$, all subsequent ones are too, so we know that $\mW^{(t)} = \w(t)\ve\ve\trs$ and the dynamics can be studied on the scalar $\w$ only.

    Because of this, all moments are given by $ \mu_k = \w^{k} ||\ve||^{2k+2} = ||\ve||^2 (\w ||\ve||^2)^{k} $. The corresponding scale and decay are respectively $||\ve||^2$ and $\w||\ve||^2$, and because of this it is only possible to obtain a Generalizing Integrator at scale $s=||\ve||^2$.

    The fixed points of the gradient descent dynamics are the real roots of the following polynomial $P$:

    \begin{equation*}
        \frac{\Delta \w}{\eta} = 2 \sum_{k=1}^{T} k (T+1-k) (\w^{k}||\ve||^{2k+2}-s\gamma^{k}) \w^{k-1}||\ve||^{2k-2} := P(\w)
    \end{equation*}

    Choosing $s=||\ve||^2$, we can check with Mathematica that for any value of $T$ larger than $2$, this polynomial has a single real root at $\w=\gamma / ||\ve||^2$, which is a generalizable minimum. Since the leading order term in this polynomial is of odd degree, we know that $\lim_{w\rightarrow \pm \infty}P(\w) = \pm \infty$, and therefore the derivative of $P$ at $\w=\gamma / ||\ve||^2$ is positive, so that this value of $\w$ is an attractive fixed-point of the dynamics. As before, the dynamics is convergent only if the learning rate is smaller than $P'(\gamma / ||\ve||^2)^{-1}$.

\section{Expression of the moments in the low rank parametrization}

\label{app:lowrank_moments}

We have defined $\mw$ so that
\begin{equation*}
\mW=\sum_{a,b=1}^{2}\w_{a,b} \overline{\vv_a}\, \overline{\vv_b}\trs.
\end{equation*}

Because of the orthogonality conditions $\overline{\vv_a}\trs\overline{\vv_b}=\delta_{a,b}$, we can easily compute the powers of $\mW$ as:
\begin{equation*}
    \mW^k = \sum_{a,b=1}^{2}(\w^k)_{a,b} \overline{\vv_a}\,\overline{\vv_b}\trs,
\end{equation*}

which yields the following moments:
\begin{equation*}
    \begin{split}
    \mu_k &= \vd\trs \mW^k \ve = \vd\trs \sum_{a,b=1}^{2}\w^k_{a,b} \overline{\vv_a}\,\overline{\vv_b}\trs \ve\\
    &= \sum_{a,b=1}^{2} \sqrt{\mSigma}_{1, a} \w^k_{a,b} \sqrt{\mSigma}_{b, 2} =  (\sqrt{\mSigma} \w^k \sqrt{\mSigma})_{1,2}\\
    \end{split}
\end{equation*}

We now assume $\mw$ to be diagonalizable as $\mw=\mP_{\w} \mLambda_{\w} \mP_{\w}^{-1}$, so that:
\begin{equation*}
    \begin{split}
        \mu_k &=   (\sqrt{\mSigma} \w^k \sqrt{\mSigma})_{1,2} = (\sqrt{\mSigma} \mP_{\w} \mLambda_{\w}^k \mP_{\w}^{-1} \sqrt{\mSigma})_{1,2}\\
        &= \sum_{i=1}^{2} \lambda_i^k (\sqrt{\mSigma} \mP_{\w})_{1, i} (\mP_{\w}^{-1} \sqrt{\mSigma})_{i,2} = \sum_{i=1}^{2} \lambda_i^k (\mP_{\w}\trs \sqrt{\mSigma})_{i, 1} (\mP_{\w}^{-1} \sqrt{\mSigma})_{i,2}\\
        &:= \sum_{i=1}^{2} g_i \lambda_i^k
    \end{split}
\end{equation*}

Using a reasoning similar to the one of Section \ref{sec:generalizing_minima}, we find the same conditions (\ref{eq:null_loss_conds}) for Generalizing Integrators, but with a new expression of the $g$ coefficients.

\section{Generalizing Integrators in the null initialization subspace}
\label{app:integrator_manifolds}
In this section, we seek to determine all matrices $\w$ that correspond to Generalizing Integrators at decay $\gamma$. A first, obvious condition is that at least one of the eigenvalues of $\w$ has to be equal to $\gamma$. Without loss of generality, we will consider this eigenvalue to be the first one, and we will denote the other $\lambda$.

The matrix $\mP_{\w}$ that diagonalizes $\mw$ can be parametrized as:
\begin{equation*}
\mP_{\w}=\begin{pmatrix}
u_1 & v_1\\
u_2 & v_2\\
\end{pmatrix}
\end{equation*}

Each column of $\mP_{\w}$ can be independently multiplied by a non-zero scalar and yield the same $\w$. There are therefore three cases: either $u_1$ or $v_1$ is null (but not both, since $\mP$ would then not be invertible), or both are non-zero.

We also recall that
\begin{equation}
    g_1 + g_2 = \sum_{i=1}^2 [(\sqrt{\Sigma}\mP_{\w})_{1, i}(\mP^{-1}_{\w}\sqrt{\Sigma})_{i, 2}]=\Sigma_{1,2}=\vd\trs\ve
\end{equation}

\subsection*{Case $u_1\neq 0$ and $v_1\neq 0$.}
In that case, the modal matrix $\mP_{\w}$ of $\w^{\bot}$ will be parametrized as follows, with $\a\neq\b$ to ensure invertibility:
\begin{equation*}
\mP_{\w}=\begin{pmatrix}
1&1\\
\a&\b\\
\end{pmatrix}
\end{equation*}

This results in the following parametrization of the space $E_{\gamma}$ of two by two matrices with at least one eigenvalue equal to $\gamma$~:
\begin{equation}\label{eq:gamma_zero_space_parametrization}
\begin{split}
E_{\gamma} &= \left \{\Gamma(\lambda, \a, \b)=\begin{pmatrix}
1&1\\
\a&\b\\
\end{pmatrix} \begin{pmatrix}
\gamma&0\\
0&\lambda\\
\end{pmatrix} \begin{pmatrix}
1&1\\
\a&\b\\
\end{pmatrix}^{-1} ; (\lambda, \a, \b)\in \mathbb{R}^3, \a\neq \b \right\}\\
&=\left \{\frac{1}{\a-\b}\begin{pmatrix}
\a\lambda-\b\gamma & \gamma-\lambda\\
\a\b(\lambda-\gamma)&\a\gamma-\b\lambda\\
\end{pmatrix} ; (\lambda, \a, \b)\in \mathbb{R}^3, \a\neq \b \right\} \\
\end{split}
\end{equation}

\paragraph{\underline{If $\lambda=\gamma$~:}}
All values of $\a$ and $\b$ correspond to $\gamma \1$, a perfect integrator at scale $s^*=\vd\trs\ve$.

\paragraph{\underline{If $\lambda\neq\gamma$ and $\lambda\neq 0$~:}}
In that case, we need to impose $g_2(\a,\b)=0$, otherwise the second eigenvalue will contribute to the output and the system will not be a Generalizing Integrator. This will in turn impose that $g_1=\vd\trs\ve$, and hence these will be integrators at scale $s=\vd\trs\ve/\gamma$. We find that:
\begin{equation*}
g_2[\a,\b] = Z \frac{(\a-\b_0)(\b-\a_0)}{\a-\b}
\end{equation*}
where:
\begin{equation*}
\begin{dcases}
\a_0&=-\frac{(\kappa-l_-)r_- + (\kappa+l_-)r_+}{2 \vd\trs\ve (r_+-r_-)} \\
\b_0&= \frac{(\kappa+l_-)r_- + (\kappa-l_-)r_+}{2 \vd\trs\ve (r_+-r_-)} \\
Z&= \frac{[\vd\trs\ve (r_+-r_-)]^2}{2 \kappa^2} \\
l_{\pm} &= ||\vd||^2\pm||\ve||^2\\
\kappa&=\sqrt{l_-^2+4 (\vd\trs\ve)^2}\in ]0, l_+[\\
r_{\pm}&=\sqrt{l_+ \pm \kappa}\\
\end{dcases}
\end{equation*}

Hence, there are two manifolds of points satisfying $g_2(\a,\b)=0$:
\begin{equation}
\begin{dcases}\label{eq:special_scale_manifolds}
\mathcal{M}_{\a} = \{\Gamma(\lambda,\a,\a_0), (\lambda,\a) \in \mathbb{R}^2\}\\
\mathcal{M}_{\b} = \{\Gamma(\lambda,\b_0,\b), (\lambda,\b) \in \mathbb{R}^2\}
\end{dcases}
\end{equation}

where $\Gamma$ is defined in equation (\ref{eq:gamma_zero_space_parametrization}). These two manifolds intersect along a $1$-Dimensional manifold:
\begin{equation*}
\mathcal{M}_{\a} \cap \mathcal{M}_{\b} = \{\Gamma(\lambda,\b_0,\a_0), \lambda \in \mathbb{R}\}
\end{equation*}

% The intersection of those manifolds with $\mathcal{M}_0$ give each a one-dimensional manifold, respectively $\{\Gamma(0,\a,\a_0), \a \in \mathbb{R}\}$ and $\{\Gamma(0,\b_0,\b), \b \in \mathbb{R}\}$, which in turn intersect at $\Gamma(0,\b_0,\a_0)$. Those two one-dimensional manifolds are exactly the ones that correspond to $s=\vd\trs\ve$ from the $\lambda=0$ case.

\paragraph{\underline{If $\lambda=0$~:}}
The system will always be a perfect integrator at decay $\gamma$, since no other eigenvalue can contribute to the output. Its scale is determined by~:
\begin{equation}\label{eq:scale_in_manifold}
s= g_1(\a,\b) = Z \frac{(\a-\a_0)(\b-\b_0)}{\b-\a}
\end{equation}

\begin{figure}
    \centering
    \includegraphics[width=1.\textwidth]{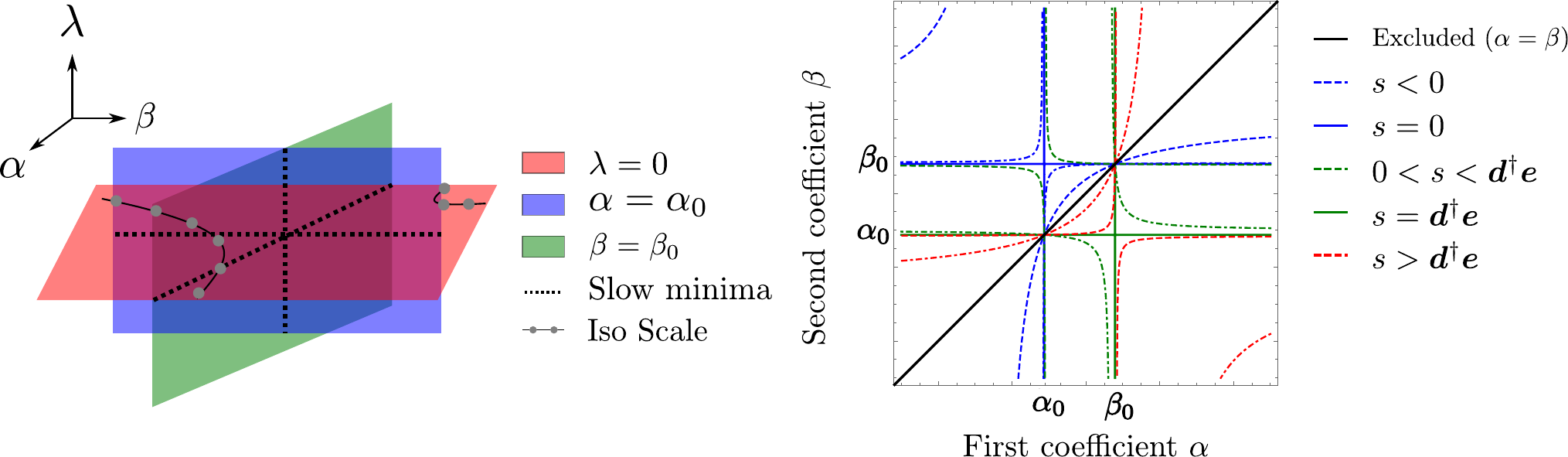}
    \caption{Structure of the minima in the null initialization subspace.  On the left, we present the structure of the manifolds of $2\times 2$--matrices with exactly one eigenvalue equal to $\gamma$, and both eigenvectors with non-zero first components. The coefficients $\a$ and $\b$ parametrize the eigenvectors, and $\lambda$ is the second eigenvalue. The "slow minima", which are located at the intersections of our manifolds, are the ones towards which convergence of Gradient Descent can be algebraically slow, see Section \ref{app:linear_special_scale}. On the right, we detail the structure of the iso-scale manifolds in the $\lambda=0$ submanifold, and show that they are indeed one-dimensional as long as $s\notin \{0, \vd\trs\ve\}$. While the lines appear to cross, they do so only on the $\a=\b$ line which is a singularity of our parametrization and therefore non-physical. }
    \label{fig:nullLambda_level_S}
\end{figure}

From this result, we deduce the following:
\begin{itemize}

\item For any given value of $s \notin \{ 0, \vd\trs\ve \}$, the set of values of $\a,\b$ that give $c_1[\a,\b]=s$ is a one-dimensional manifold, as can be seen in Figure \ref{fig:nullLambda_level_S}. A parametrization of this manifold can be obtained by inverting equation (\ref{eq:scale_in_manifold}) as the set $\Gamma(0, \a, \b_s(\a))$ where $\Gamma$ is defined in equation (\ref{eq:gamma_zero_space_parametrization}) and:
\begin{equation*}
    \b_s(\a) = \frac{Z(\a-\a_0)\b_0  - \a s }{Z(\a-\a_0)-s}.
\end{equation*}

\item When $s=0$, the two solutions are $\a=\a_0$ or $\b=\b_0$, no matter the value of the other parameter.

\item When $s=\vd\trs\ve$, equation (\ref{eq:scale_in_manifold}) is not invertible and the condition $g_1=s$ is satisfied if and only if $\b=\a_0$ or $\a=\b_0$, which are exactly the intersection of the $\mathcal{M}_{\a}$ (resp. $\mathcal{M}_{\b}$) manifolds of equation (\ref{eq:special_scale_manifolds}) with the $\lambda=0$ subspace.
\end{itemize}

\subsection*{From $\w$ to $\mW$}
We have shown that in the generic case $s\notin \{ 0, \vd\trs\ve\}$, the Generalizing Integrators at scale $s$ correspond to $\w$ of rank 1 and form a $1$--dimensional manifold.

Such matrices $\w$ can be parametrized as:
\begin{equation*}
\begin{split}
\mathcal{M}_{\lambda=0}&= \left \{\frac{\gamma}{\b-\a}\begin{pmatrix}
\b & -1\\
\a\b &-\a\\
\end{pmatrix} ; (\a, \b)\in \mathbb{R}^2, \a\neq \b\right\}\\
&=\left \{\sigma_{\w} \vx \vy\trs;\sigma_{\w} = \frac{\gamma \sqrt{(\a^2+1)(\b^2+1)}}{\b-\a}, \vx=\frac{1}{\sqrt{\a^2+1}} (1, \a), \vy=\frac{1}{\sqrt{\b^2+1}}  (\b, -1), (\a, \b)\in \mathbb{R}^2, \a\neq \b\right \}
\end{split}
\end{equation*}
where $\vx$ and $\vy$ are respectively the left and right eigenvector of the corresponding rank $1$ matrix. When $\w$ is of that form, we have:
\begin{equation*}
    \mW=\sigma_{\w} \sum_{a,b}\vx_a \vy_b \overline{\vv_a} \, \overline{\vv_b}\trs = \sigma_{\w} (\sum_{a}\vx_a \overline{\vv_a})(\sum_{b}\vy_b \overline{\vv_b}\trs):= \sigma \vl \vr\trs,
\end{equation*}
so that $\mW(\w)$ is of rank 1 too.

% One can also note that:
% \begin{equation*}
%     \lim_{x\rightarrow\pm\infty} \b_s(\a) = 1-s/Z
% \end{equation*}
% which leads to:
% \begin{equation*}
%     \lim_{x\rightarrow\pm\infty} E_{\gamma}(\lambda=0, \a, \b_s(\a)) = \begin{pmatrix}
% 0 & 0\\
% \gamma(1-s/Z) &\gamma\\
% \end{pmatrix}
% \end{equation*}

% meaning that the large $\a$ limits correspond to finite $\mw$

\subsection*{Case $u_1 = 0$}
The matrix $\mP_{\w}$ that diagonalizes $\w$ will be parametrized as:
\begin{equation*}
\mP_{\w}=\begin{pmatrix}
0&1\\
1&v\\
\end{pmatrix}
\end{equation*}
and is always invertible no matter the value of $v\neq 0$.

Now, there are two degrees of freedom $\lambda$ and $v$, and the corresponding matrices are:
\begin{equation*}
\mw=\begin{pmatrix}
\lambda & 0\\
(\gamma-\lambda)v &\gamma\\
\end{pmatrix}
\end{equation*}

As before, the scale $s$ is determined by $c_{\gamma}$, which depends linearly on $v$. Hence, the choice of scale fixes $v$, and either $\lambda$ has to be chosen either equal to $0$ when $s$ is different from $\vd\trs\ve$ (yielding a single solution) or it remains free if the choice of scale imposes $c_{\lambda}=0$ (yielding a $1D$ manifold).

A perfectly analogous reasoning can be applied when $v_1=0$ and $u_1\neq 0$, and in both cases the manifolds of solution are of lower dimension than their counterparts which have non-zero $u_1$ and $u_2$; we discard those solutions as we expect them to be smooth limits of the generic case.

\section{Gradients and Hessian in the low-rank parametrization}
\label{app:linear_lowrank_hessian}
    We will now explicitly compute the derivative of the loss with respect to the $2\times 2$--matrix $\w$. We previously found that:
    \begin{equation*}
    \begin{split}
        \mu_q &= \vd\trs\mW^q\ve = (\sqrt{\mSigma}\, \mw^{q}\, \sqrt{\mSigma})_{12} \\
        &= \sum_{a,b=1}^2 \sqrt{\mSigma}_{1a} \mw^{q}_{ab} \sqrt{\mSigma}_{b2}
    \end{split}
    \end{equation*}

    and we have that:
    \begin{equation*}
        \frac{\partial \mw^{q}_{a,b}}{\partial \w_{ij}} = \sum_{m=0}^{q-1}\mw^{m}_{ai}\mw^{q-1-m}_{jb}
    \end{equation*}

    so that:
    \begin{equation*}
        \frac{\partial \mu_q}{\partial \w_{ij}} = \sum_{m=0}^{q-1} (\sqrt{\mSigma}\w^{m})_{1i} (\w^{q-1-m} \sqrt{\mSigma})_{j2}
    \end{equation*}

    which allows us to compute the gradient of the loss with respect to the coefficients of $\mw$ through:
    \begin{equation*}
        \frac{\partial \mathcal{L}}{\partial \w_{ij}} = 2 \sum_{q,p=1}^T \chi_{q,p}(\mu_q -s \gamma^q) \frac{\partial \mu_p}{\partial \w_{ij}}
    \end{equation*}

    We can now compute the Hessian of the loss, which will allow us to derive formulas for stability and speed of convergence of Gradient Descent. This Hessian can be decomposed as follows:
    \begin{equation*}
       \mathcal{H}_{ij,kl} = \frac{\partial \mathcal{L}}{\partial \w_{ij}\partial \w_{kl}} = \sum_{q,p=1}^T \chi_{q,p}\left [\frac{\partial \mu_q}{\partial \w_{ij}} \frac{\partial \mu_p}{\partial \w_{kl}}+(\mu_q -s \gamma^q)\frac{\partial \mu_p}{\partial \w_{ij}\partial \w_{kl}} \right]
    \end{equation*}

    % \begin{equation*}
    %     \frac{\partial \mathcal{L}}{\partial \w_{ij}\partial \w_{kl}} = \sum_{q=1}^T (T+1-q)[\frac{\partial \mu_q}{\partial \w_{ij}} \frac{\partial \mu_q}{\partial \w_{kl}}+(\mu_q -s \gamma^q)\frac{\partial \mu_q}{\partial \w_{ij}\partial \w_{kl}}]
    % \end{equation*}

    We want to estimate this Hessian at rank 1 generalizing integrators, so that the second part will always be zero. Therefore, the Hessian is simply:

    % \begin{equation*}
    %     \frac{\partial \mathcal{L}}{\partial \w_{ij}\partial \w_{kl}} = \sum_{q=1}^T (T+1-q) \left [\sum_{m=0}^{q-1} (\sqrt{\mSigma}\mw^{m})_{1i} (\mw^{q-1-m} \sqrt{\mSigma})_{j2}\right ]
    %     \left [\sum_{\Tilde{m}=0}^{q-1} (\sqrt{\mSigma}\mw^{\Tilde{m}})_{1k} (\mw^{q-1-\Tilde{m}} \sqrt{\mSigma})_{l2} \right]
    % \end{equation*}

    \begin{equation}\label{app:lowrank_hessian}
        \mathcal{H}_{ij,kl} = \sum_{q,p=1}^T \chi_{q,p} \left [\sum_{m=0}^{q-1} (\sqrt{\mSigma}\mw^{m})_{1i} (\mw^{q-1-m} \sqrt{\mSigma})_{j2}\right ]
        \left [\sum_{\Tilde{m}=0}^{p-1} (\sqrt{\mSigma}\mw^{\Tilde{m}})_{1k} (\mw^{p-1-\Tilde{m}} \sqrt{\mSigma})_{l2} \right]
    \end{equation}

\section{Minimum convergence time}
\label{app:convergence_time}
We are now interested in studying the dynamics of convergence towards the GIs $\mW^*$ in the null initialization subspace. To do so, we use a Taylor expansion of the loss around one of its minima:

\begin{equation*}
    \mathcal{L}(\mW^*+\delta \mW) = \sum_{i,j,k,l=1}^n \delta \mW_{ij} \; \mathcal{H}_{ij,kl}(\mW^*)\; \delta \mW_{kl}
\end{equation*}

Seeing $\delta\mW$ as a vector and $\mathcal{H}$ as a (symmetric) matrix, we can diagonalize $\mathcal{H}$ with real eigenvalues $\lambda_I$ and normalized eigenvectors $\vu_I$, and express $\delta\mW=\sum_{I=1}^{n^2} \delta_I \vu_I$ in that basis so that:
\begin{equation}
    \mathcal{L}(\mW^*+\delta \mW) = \sum_{I=1}^{n^2} \lambda_I \, \delta_I^2
\end{equation}

Since our loss is positive for any weight-matrix $\mW$, we expect that all eigenvalues of the Hessian computed at a GI be positive. We also expect that (in the generic case $s\neq \vd\trs\ve$) one of them is $0$, corresponding to the local tangent to the $1$--dimensional manifold of GIs.

Writing the $GD$ dynamics on the perturbation $\vdelta$, we find that:
\begin{equation*}
    \delta_I^{(\tau+1)} = (1 - \eta \lambda_I) \delta_I^{(\tau)},
\end{equation*}
hence $\vdelta$ will see each of its components either be conserved (if it corresponds to a null eigenvalue) or evolve exponentially. This exponential evolution is convergent as long as $|1 - \eta \lambda_I|<1$, and monotonic as long as $\eta < 1/\lambda_I$. Choosing the optimal learning rate for the full system $\eta^*=1/\lambda_{\max}$, the slowest component of $\vdelta$ evolves as
$$(1 - \eta^* \lambda_{min})^{\tau}=(1 - \lambda_{min}/\lambda_{max})^{\tau}\simeq e^{\tau \ln{(1-\mathcal{C}^{-1})}}\simeq e^{-\tau/\mathcal{C}},$$ hence the characteristic convergence time will be $\mathcal{C}=\lambda_{max}/\lambda_{min}$\footnote{ $\lambda_{min}$ is the minimum \textbf{non-zero} eigenvalue.}

\begin{figure}
    \centering
    \includegraphics[width=\textwidth]{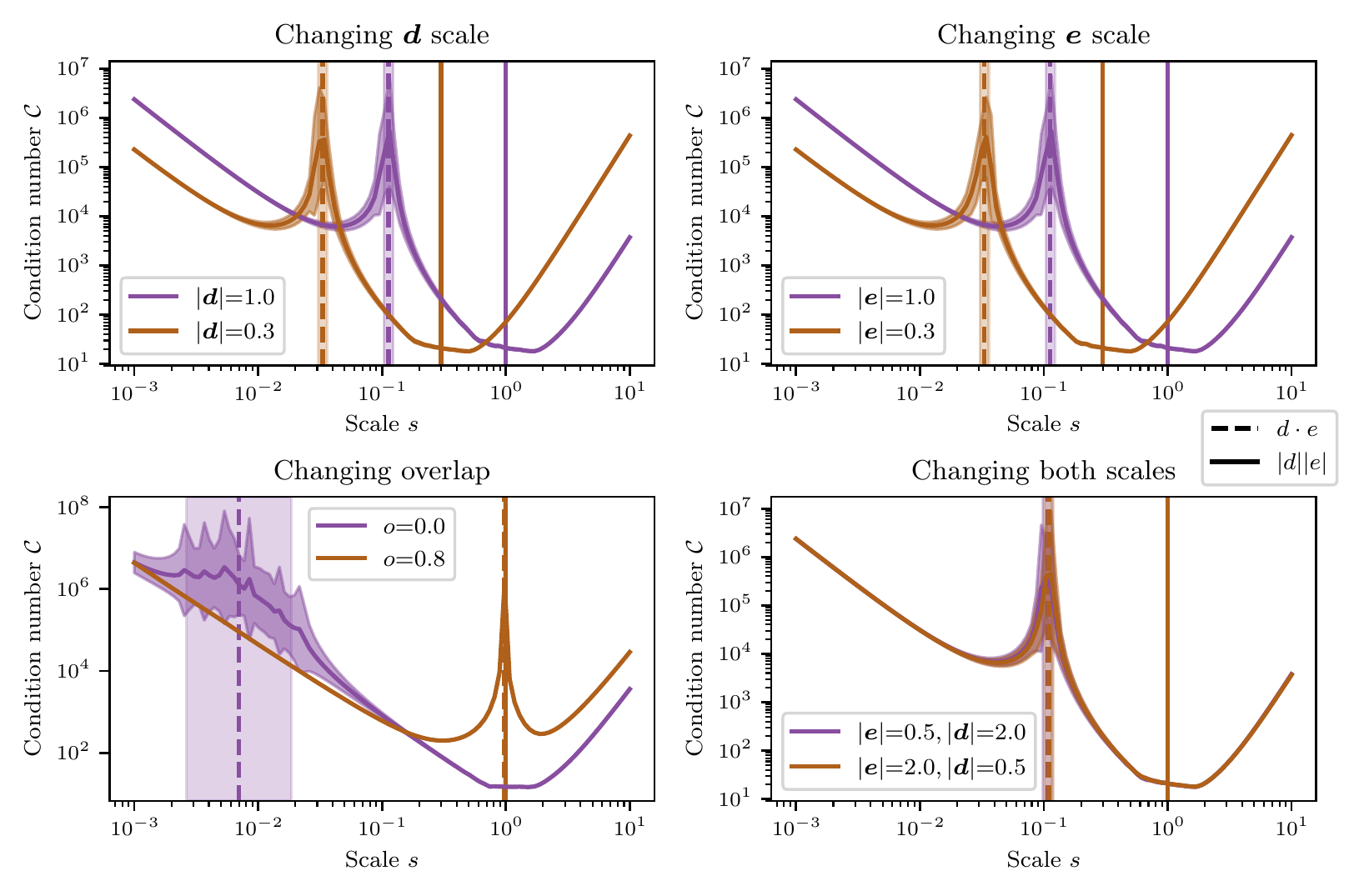}
    \caption{Evolution of the minimum convergence time as a function of the scale.} \label{fig:full_condition_linear}
\end{figure}

Since in the null initialization case the weights are parametrized by a $2\times 2$--matrix, the Hessian is $4\times 4$ and its spectrum can easily be computed numerically by using equation (\ref{app:lowrank_hessian}). We therefore performed the following study: fixing the $L_2$ norm of the encoder and decoder as well as their dot product\footnote{ In order to generate $\ve$ and $\vd$ with macroscopic overlaps (larger than $n^{-1/2}$), we first draw them independently, normalize them to 1, then modify the decoder as $\vd=o \ve + (1-o)\vd$ where $o$ is the overlap; for large enough $n$, the dot product $\vd\trs\ve$ will be close to this overlap. We then independently rescale them to the desired norm.}, we evaluate the spectrum of $\mathcal{H}$ and deduce from it the condition number $\mathcal{C}$  along each manifold of rank 1 $s$--scaled GIs; we find that a minimum exists for $\a,\b$ (see Appendix \ref{app:integrator_manifolds}) of order 1, and this value will be a lower bound for the convergence time to \textbf{any} GI at scale $s$ using Gradient Descent. We plot the value of this bound as a function of $s$ for different initial choices of i/o--vectors in Figure \ref{fig:full_condition_linear}.

\section{Algebraic convergence for specific scale value }
\label{app:linear_special_scale}

\begin{figure}
        \includegraphics[width=\textwidth]{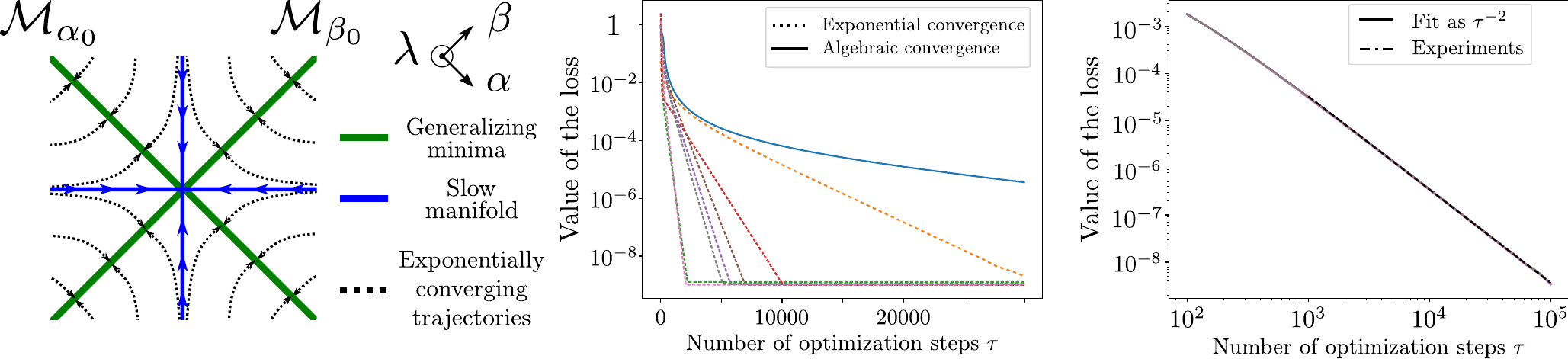}
        \caption{Explanation of the dynamics of convergence towards a minimum at $s=\vd\trs\ve$. (Left) Informal representation of GD trajectories. Two types of trajectories converging to GIs exist: starting from a random $\mW$ in the null initialization subspace (middle), we usually converge outside of the intersection of the two manifolds exponentially fast, with a fairly wide range of times of convergence depending on the precise starting point; in rare cases, that random starting point lies on (or close enough to) the slow manifold on which convergence is as a power law $\tau^{-2}$. We illustrate this algebraic convergence by starting from $\mW=0$ (right), which is experimentally found to be on the slow manifold. Both experimental curves show $8$ different realizations of the training, with same learning rate, norms of vectors and overlap (but scale chosen exactly to $\vd\trs\ve$ after $\ve$ and $\vd$ have been drawn for that particular realization of the experiment). \label{fig:linear_exp_vs_alg}}
    \end{figure}

    From the previous analysis, it seems that when $s=\vd\trs\ve$ the $\lambda=0$ manifold of solutions is hard to reach. Numerically, we see that Gradient Descent converges to a solution that lies in the union of the two $2D$ manifolds described earlier, corresponding to $\mW$ of rank a priori $2$. If we initialize with a random, non-zero, $\mW$ in the null initialization subspace, we converge exponentially; if we start from $\mW=0$, the convergence is algebraic as a power law $\tau^{-2}$  instead of exponential (see Figure \ref{fig:linear_exp_vs_alg}).

    To understand this phenomenon, we will consider the continuous-time, non-linear differential equation on the coefficients of $\mw$: $\dot{\mw}=-\partial_{\mw} \mathcal{L}$. Let us also introduce the two manifolds of GIs at scale $s=\vd\trs\ve$:

    \begin{equation*}
    \begin{split}
    &\mathcal{M}_{\a_0}(\a, \lambda)=\frac{1}{\a-\a_0}
        \begin{pmatrix}
        \a \lambda - \a_0 \gamma & \gamma-\lambda\\
        \a \a_0 (\lambda-\gamma) & \a \gamma-\a_0 \lambda\\
        \end{pmatrix} \hspace{0.4cm} \\
        & \mathcal{M}_{\b_0}(\b, \lambda)=\frac{1}{\b-\b_0}
        \begin{pmatrix}
        \b \gamma - \b_0 \lambda & \lambda-\gamma\\
        \b \b_0 (\gamma-\lambda) & \b\lambda- \a_0\gamma \\
        \end{pmatrix}
    \end{split}
    \end{equation*}

    In the following, we refer respectively to the union and the intersection of those manifolds as $\mathcal{U}$ and $\mathcal{I}$. If we consider any GI $\mM$ in $\mathcal{U}\setminus \mathcal{I}$, numerical experiments show that the Hessian has exactly two null and two strictly positive eigenvalues; the two null directions, which give us the linearized centre space $E_c$ around $\mM$ in which convergence is at most as a power law, correspond to the local tangent of the manifold of minima and are therefore not relevant: convergence of the loss is exponential.

    On the other hand, if $\mM\in\mathcal{I}$, the Hessian exhibits three null eigenvalues (because the manifolds of minima intersect non-tangentially), so that the centre space $E_c$ is now of dimension $3$. Since the GIs are only two $2D$ planes, there exists an invariant manifold along which convergence is not exponential.
    Denoting as $x$ the coordinate along that slow direction, the Center Manifold Theorem ensures that the evolution of $x$ is given by:
    \begin{equation*}
        \dot{x} = g(x),
    \end{equation*}
    where $g$ is a polynomial of order at least 2 with neither constant nor first order term.

    Assuming that the order two term is non-zero, we get that locally, for $x$ close to $0$, $\dot{x}=a x^2$. Integrating over time, we get that $x$ evolves as $\tau^{-1}$.
    We then look at the value of the loss when $\mw$ is equal to a generalizing minimum $\mM$ plus a small perturbation $\mX$ of order $x$:
    % \begin{equation*}
    %     \begin{split}
    %         \mathcal{L}&=\sum_{q=1}^T(\mu_q-s\gamma^q)^2 =\sum_{q=1}^T [\sqrt{\mSigma}\, (\mM+\mX)^{q}\, \sqrt{\mSigma})_{12}-s\gamma^q]^2\\
    %         &= \sum_{q=1}^T [(\sqrt{\mSigma}\, \mM^{q}\, \sqrt{\mSigma})_{12}-s\gamma^q + \mathcal{O}(x)]^2 \\
    %         &= \mathcal{O}(x^2)
    %     \end{split}
    % \end{equation*}

    \begin{equation*}
        \begin{split}
            \mathcal{L}&=\sum_{q,p=1}^T\chi_{qp}(\mu_q-s\gamma^q)(\mu_p-s\gamma^p) =\sum_{q,p=1}^T\chi_{qp} [\sqrt{\mSigma}\, (\mM+\mX)^{q}\, \sqrt{\mSigma})_{12}-s\gamma^q][\sqrt{\mSigma}\, (\mM+\mX)^{p}\, \sqrt{\mSigma})_{12}-s\gamma^p]\\
            &= \sum_{q,p=1}^T\chi_{qp} [(\sqrt{\mSigma}\, \mM^{q}\, \sqrt{\mSigma})_{12}-s\gamma^q + \mathcal{O}(x)][(\sqrt{\mSigma}\, \mM^{p}\, \sqrt{\mSigma})_{12}-s\gamma^p + \mathcal{O}(x)] \\
            &= \mathcal{O}(x^2)
        \end{split}
    \end{equation*}

    Therefore, if the quadratic term of $g$ is non-zero, $x$ scales as $\tau^{-1}$ and the loss as $\tau^{-2}$ when $\tau$ is large, as is observed experimentally. It should be noted that this result does not depend on the value of $T$ nor on the choice of $\ve$ and $\vd$, as observed experimentally too.

    Therefore, algebraic convergence is observed only when very strict conditions are met:
    \begin{itemize}
        \item the gradient descent starts from a very specific subspace, the pre-image of the intersection, which we will refer to as a "slow manifold". It is of lower dimension than the initial space of weight-matrices, so that random initial conditions will almost never satisfy this criterion.
        \item the system always remains on the trajectory of GD. In particular, this means that $\eta$ and the noise on the computed updates need to be small enough that we don't accidentally leave the slow manifold, which would then lead to exponential convergence.
    \end{itemize}

\section{Single-channel ReLU proxy loss gradients and Hessian}
\label{app:relu_proxy_gradients}
We have shown in Section \ref{sec:relu_proxy} that the two following pairs of conditions are enough to guarantee perfect integration of arbitrary signals:
\begin{equation}
    \begin{cases}
     \vd\trs \R{\pm \mW \ve} &= \pm s\gamma \\
     \mW \R{\pm \mW \ve} &= \pm \gamma \mW \ve \\
    \end{cases}
\end{equation}

We define the \newterm{proxy} loss as the sum of four terms corresponding to the residuals of those equalities:
\begin{equation*}
    \mathcal{L}^{proxy} = \mathcal{L}^1_+ + \mathcal{L}^1_- + \mathcal{L}^2_+ + \mathcal{L}^2_-
\end{equation*}
where $\mathcal{L}^1_{\pm}=(\vd\trs \R{\pm \mW \ve} - \, \pm s \gamma)^2$ and $\mathcal{L}^2_{\pm}=|\mW \R{\pm \mW \ve} - \, \pm  \gamma \mW \ve|^2$.

{\flushleft The gradients of these quantities are computed as~:}

\begin{equation*}
\begin{dcases}
\frac{\partial \mathcal{L}^1_{\pm}}{\partial \mW_{ij}} &= 2(\vd\trs \R{\pm \mW \ve} - \, \pm s \gamma)(\pm \vd_i \H(\pm\mW\ve)_i \ve_j) \\
\frac{\partial \mathcal{L}^2_{\pm}}{\partial \mW_{ij}} &= 2 (\mW \R{\pm \mW \ve} - \, \pm  \gamma \mW \ve)|_i(\R{\pm \mW \ve} - \pm s\gamma)|_j \pm 2 e_j \sum_a (\mW \R{\pm \mW \ve} - \, \pm  \gamma \mW \ve)|_a \mW_{ai} \H(\pm \mW\ve)|_i\\
\end{dcases}
\end{equation*}

where $\H$ is the componentwise Heaviside function, that takes a vector as input and returns a vector whose $k$-th component is $0$ if the $k$-th component of the input was strictly negative, $1$ if it was strictly positive, and $.5$ if it is exactly $0$.

The Hessian of the $\mathcal{L}^1$ terms is readily computed as~:

\begin{equation*}
\frac{\partial \mathcal{L}^1_{\pm}}{\partial \mW_{ij}\partial \mW_{kl}} = 2 \vd_i \H(\pm \mW\ve)|_i \ve_j \vd_k \H(\pm \mW\ve)|_k \ve_l) + 2(\vd\trs \R{\pm \mW \ve} - \, \pm s \gamma)\delta_{ik} d_i e_j e_l \vdelta(\pm \mW\ve)|_i
\end{equation*}

where $\delta$ is the discrete Dirac distribution $\delta_{ik}$ which is one if $i=k$ and $0$ otherwise, and $\vdelta$ the componentwise Dirac distribution such that $\vdelta(\vv)$ is a vector of same shape as $\vv$ whose components are $1$ if the corresponding composant of $\vv$ is $0$, and $0$ otherwise.
This part of the Hessian is indeed symmetric by exchange of $ij$ and $kl$ because of the $\delta_{ik}$ in the second term.

The Hessian of the $\mathcal{L}^2$ terms is more complicated, but can be found to be~:

\begin{equation*}
\begin{split}
\frac{1}{2}\frac{\partial \mathcal{L}^2_{\pm}}{\partial \mW_{ij}\partial \mW_{kl}} &=
\sum_a \mW_{ai}\mW_{ak} e_j e_l \H(\pm\mW\ve)_i \H(\pm\mW\ve)_k \\
&+ \sum_a (\mW \R{\pm \mW \ve} - \, \pm  \gamma \mW \ve)|_a \mW_{ai} \delta_{ik}  \vdelta(\pm\mW\ve)|_i  e_j e_l\\
&+ \delta_{ik} (\R{\pm \mW \ve} - \, \pm  \gamma \ve)|_j \, \, (\R{\pm \mW \ve} - \, \pm  \gamma \ve)|_l\\
&+ \pm[\delta_{jk}(\mW \R{\pm \mW \ve} - \, \pm  \gamma \mW \ve)|_i e_l \H(\pm\mW\ve) \text{ + $ij \Leftrightarrow kl$}]\\
&+ \pm[(\R{\pm \mW \ve} - \, \pm  \gamma \ve)|_j \mW_{ik} e_l \H(\pm\mW\ve)|_k \text{ + $ij \Leftrightarrow kl$}]\\
\end{split}
\end{equation*}

Combining those four terms, we get the Hessian of our full proxy loss:

\begin{equation*}
\begin{split}
\frac{1}{2}\frac{\partial \mathcal{L}^{proxy}_{\pm}}{\partial \mW_{ij}\partial \mW_{kl}} &= d_i e_j d_k e_l [\H(\mW \ve)|_i\H(\mW \ve)|_k+\H(-\mW \ve)|_i \H(-\mW \ve)|_k]\\
&+ e_j e_l \sum_a \mW_{ai}\mW_{ak}  [\H(\mW \ve)|_i\H(\mW \ve)|_k+\H(-\mW \ve)|_i \H(-\mW \ve)|_k] \\
&+ \delta_{ik} [(\R{\mW \ve} - \gamma \ve)|_j(\R{\mW \ve} -  \gamma \ve)|_l + (\R{- \mW \ve} +  \gamma \ve)|_j (\R{-\mW \ve} + \gamma \ve)|_j] \\
&+ [\delta_{jk} (\mW^2\ve-2\gamma\mW\ve)|_i e_l \text{ + $ij \Leftrightarrow kl$}] \\
&+ [\mW_{ik} (\mW\ve-2\gamma\ve)|_j e_l \text{ + $ij \Leftrightarrow kl$}] \\
&+ \delta_{ik} d_i e_j e_l \vd\trs|\mW\ve| \vdelta(\mW\ve)|i\\
&+ \delta_{ik} e_j e_l \vd\trs|\mW\ve| \sum_a \mW_{ai}(\mW|\mW\ve|)|_a   \vdelta(\mW\ve)|i\\
\end{split}
\end{equation*}

\begin{figure}
\centering
    \includegraphics[width=.8\textwidth]{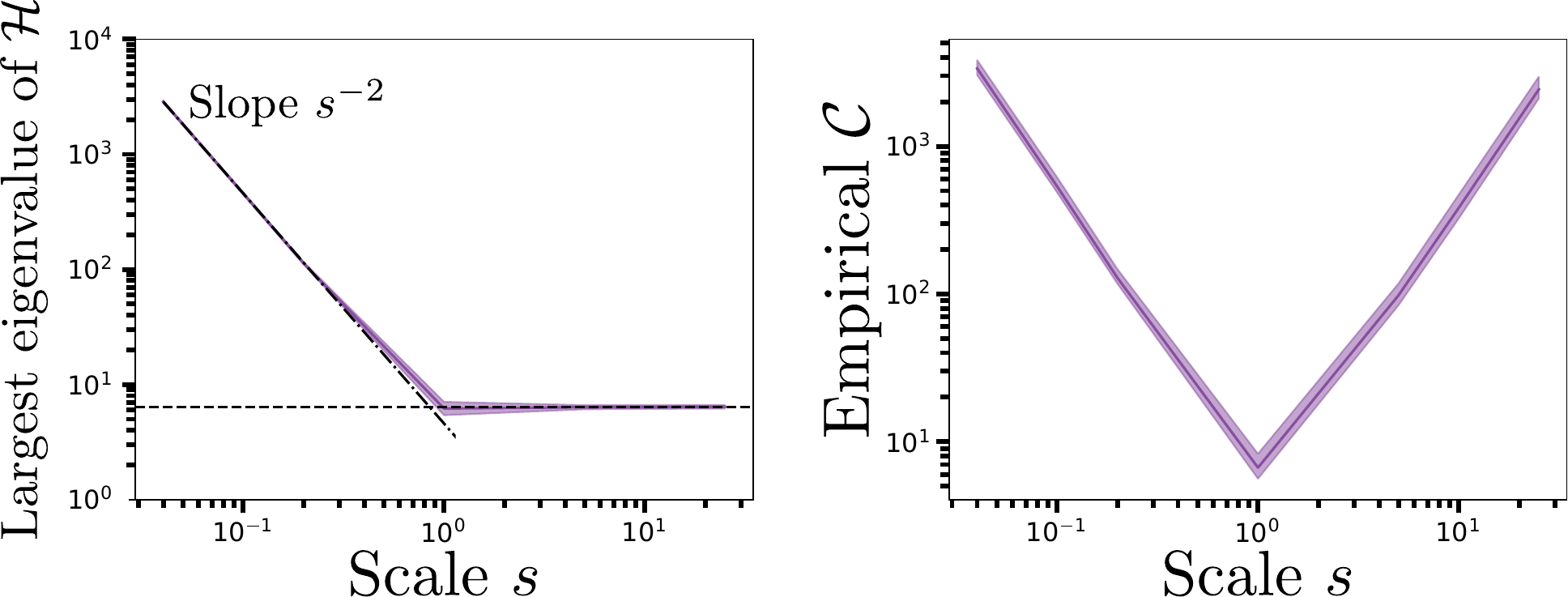}
    \caption{Experimentally determined values of the highest eigenvalue of the Hessian around the GI manifold, determining the optimal stable learning rate for GD, and of the empirical convergence time as a function of the scale. Numerical experiments carried out with $n=50$ neurons, independent encoder and decoder of norm $1$.   \label{fig:relu_CN}}
\end{figure}

Contrary to the linear null initialization case where the Hessian was a $4\times 4$--matrix, we have no way to a priori reduce the number of degrees of freedom, and $\mathcal{H}$ is a $n^2\times n^2$--matrix. We are therefore restricted to very low number of neurons (around 50 in our case) for the diagonalization to remain computationally tractable.
Another major obstacle is that we do not have an analytical expression of the GI manifolds at which we want to evaluate the Hessian. We adopted the following methodology: first, we train networks on the proxy loss using Gradient Descent at low learning rate and wait for convergence; we evaluate the largest eigenvalue $\lambda_+$ of $\mathcal{H}$ at the obtained weight-matrix, but do not compute the lowest ones as they are both prone to numerical errors, and not necessarily positive as some small negative eigenvalues will exist when we are only close to a GI; we compute an "effective" lowest eigenvalue by fitting the decay of the loss during GD at learning rate $\eta<1/\lambda_+$, and deduce the corresponding minimum convergence time. The results of these numerical simulations are presented in Figure \ref{fig:relu_CN}. We performed tests on larger networks to verify if the inferred maximum stable learning rate remained valid, as well as the order of magnitude of the convergence time, yielding the expected results.

\section{Analysis of rank-1 ReLU generalizing integrators}
\label{app:rank_1_relu}

\begin{figure}[h!]
\centering
    \includegraphics[width=0.8\textwidth]{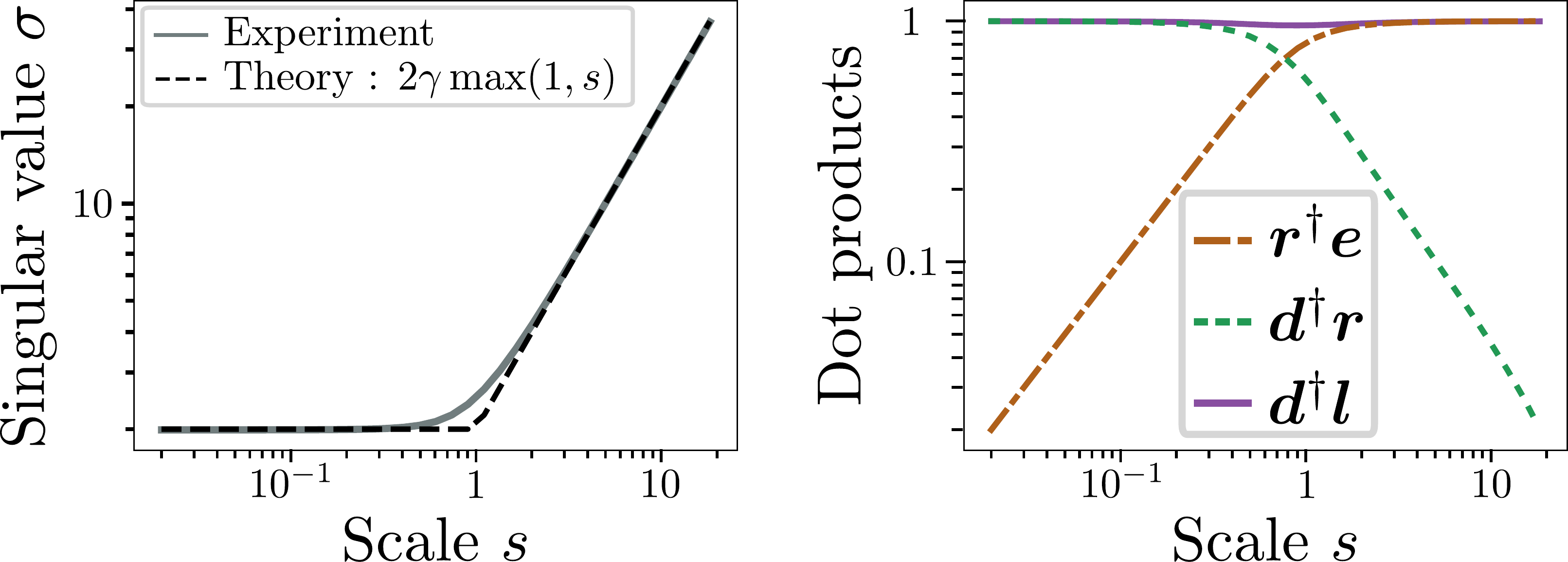}
    \caption{Behavior of the singular value $\sigma$ and dot products of the singular vectors $\vl,\vr$ with $\ve,\vd$ as functions of $s$. The figure was obtained with $n=1000$, independently drawn encoder and decoder, and aggregated across $6$ realizations of GD on the proxy loss. Error bars are not reported as they are not distinguishable from the line width. \label{fig:understand_relu_solution}}
\end{figure}

Training of the ReLU RNNs, either on real data or on the proxy loss (\ref{eq:proxy_loss_relu}), leads to GIs exhibiting one dominant singular value. As in the linear case, we write $\mW=\sigma\vl \vr\trs$; with no loss of generality we can impose $\vr\trs\ve>0$  by multiplying $\vr$ and $\vl$ by $-1$. Conditions (\ref{condloss}) become
\begin{equation*}
\begin{cases}
\sigma \, \vr\trs\vl_{\pm} &= \pm \gamma\\
\sigma (\vr\trs\ve)(\vd\trs\vl_{\pm}) &= \pm s \gamma\\
\end{cases}
\text{ where } \vl_{\pm} = \R{\pm \vl}.
\end{equation*}
%In the rest of this section, we assume these conditions to be satisfied and derive their geometrical interpretation.
Using Cauchy-Schwarz inequality, and denoting as $\bm{1}_{\pm}$ the vector whose component $i$ is equal to 1 if $\vl_i$ is of the corresponding sign and $0$ otherwise, we have:
\begin{equation*}
    |\vd\trs\vl_{\pm}|=|(\vd \bm{1}_{\pm})\trs(\pm \vl \bm{1}_{\pm})|\leq|\vd \bm{1}_{\pm}||\vl\bm{1}_{\pm}|.
\end{equation*}
Assuming half of the components of $\vl$ are positive and half are negative, as confirmed by numerical studies, and given that $|\vd|=|\vl|=1$, both norms on the right hand side are equal to $2^{-1/2}$ in the large $n$ limit, yielding $|\vd\trs\vl_{\pm}|\leq 1/2$. Since $0\leq\vr\trs\ve\leq 1$, we conclude that $\sigma\geq 2 s \gamma$. Similarly, we have that $|\vr\trs\vl_{\pm}|\leq 1/2$, implying $\sigma\geq 2 \gamma$. These conditions can then be summarized into $\sigma \geq 2 \gamma \max(s, 1)$.

Experimentally, we find that this lower bound is closely followed when $s$ is either large or small. Since $\mW$ is of rank 1, its Frobenius norm is equal to $\sigma$, and we argue that the saturation of this lower-bound on $\sigma$ is a manifestation of the conjecture of \citep{ImplicitRegularization} that gradient descent implicitly favors solutions with small matrix norm. Therefore, we have for any scale $s$ significantly different from 1:
\begin{equation}
    \sigma = 2 \gamma \max(s, 1).
\end{equation}
% which is illustrated in Figure \ref{fig:understand_relu_solution}.

\begin{figure}[h!]
\centering
    \includegraphics[width=0.8\textwidth]{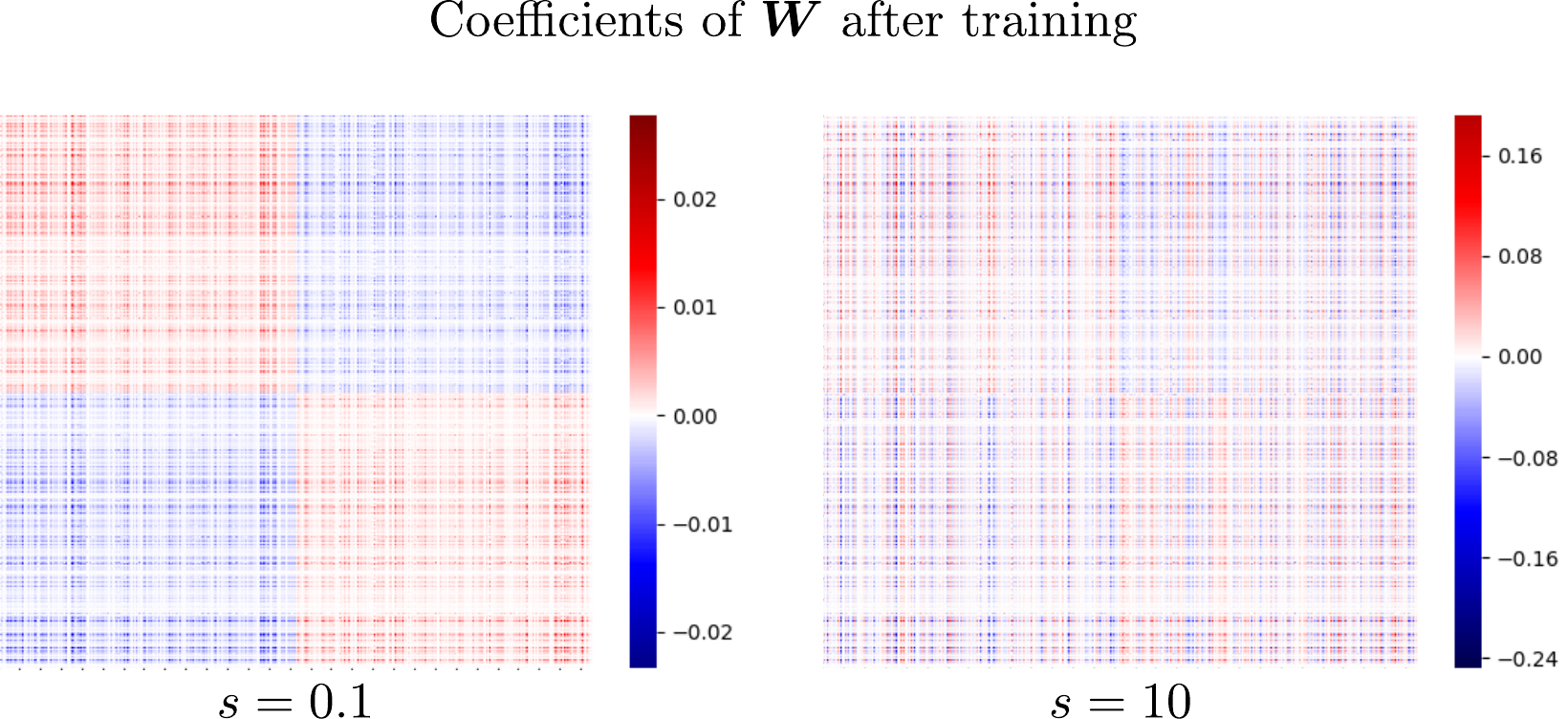}
    \caption{Weight-matrix $\mW$ of a single-channel ReLU network visualized as a discrete heatmap. Neurons were reordered so that the indices of the "+" population are from $0$ to $n/2$, while the "-" population goes from $n/2$ to $n$. When $s=0.1$, the sign of $W_{ij}$ is fully determined by whether $i$ and $j$ are in the same cluster; when $s=10$, this observation is no longer true. We also note that, as expected, the coefficients of $\mW$ are larger when $s$ increases as the norm of $\mW$ scales as $\max(1,s)$. This figure was obtained for independent encoder and decoder of scale $1$, $n=1000$. \label{fig:clustered_W}}
\end{figure}

Numerical experiments show that, for a wide range of scales, $\vl$ and $\vd$ are almost equal. Hence, $\vd\trs\vl_{\pm} = \pm |\vd\R{\pm\vd}|^2\simeq \pm 1/2$, entailing $\vr\trs\ve \simeq \min(s, 1)$ . For $s\ll 1$, $\vr$ is almost aligned with $\vd$, while for $s>1$ we have $\vr\simeq\ve$ (This statement holds for uncorrelated encoder and decoder). Our theoretical predictions are in very good agreement with numerical experiments, as shown in Figure \ref{fig:understand_relu_solution}. Notice that the change of direction of $\vr$ with $s$ has consequences on the signs of the couplings: $W_{ij}$ is positive for pairs of neurons within the "+" and "-" populations and negative in between at small $s$, but is essentially random at large $s$, see Figure \ref{fig:clustered_W}.

\section{Example of transfer learning: context-dependent selectivity}
\label{app:mante}

\begin{figure}[h!]
\centering
    \includegraphics[width=0.7\textwidth]{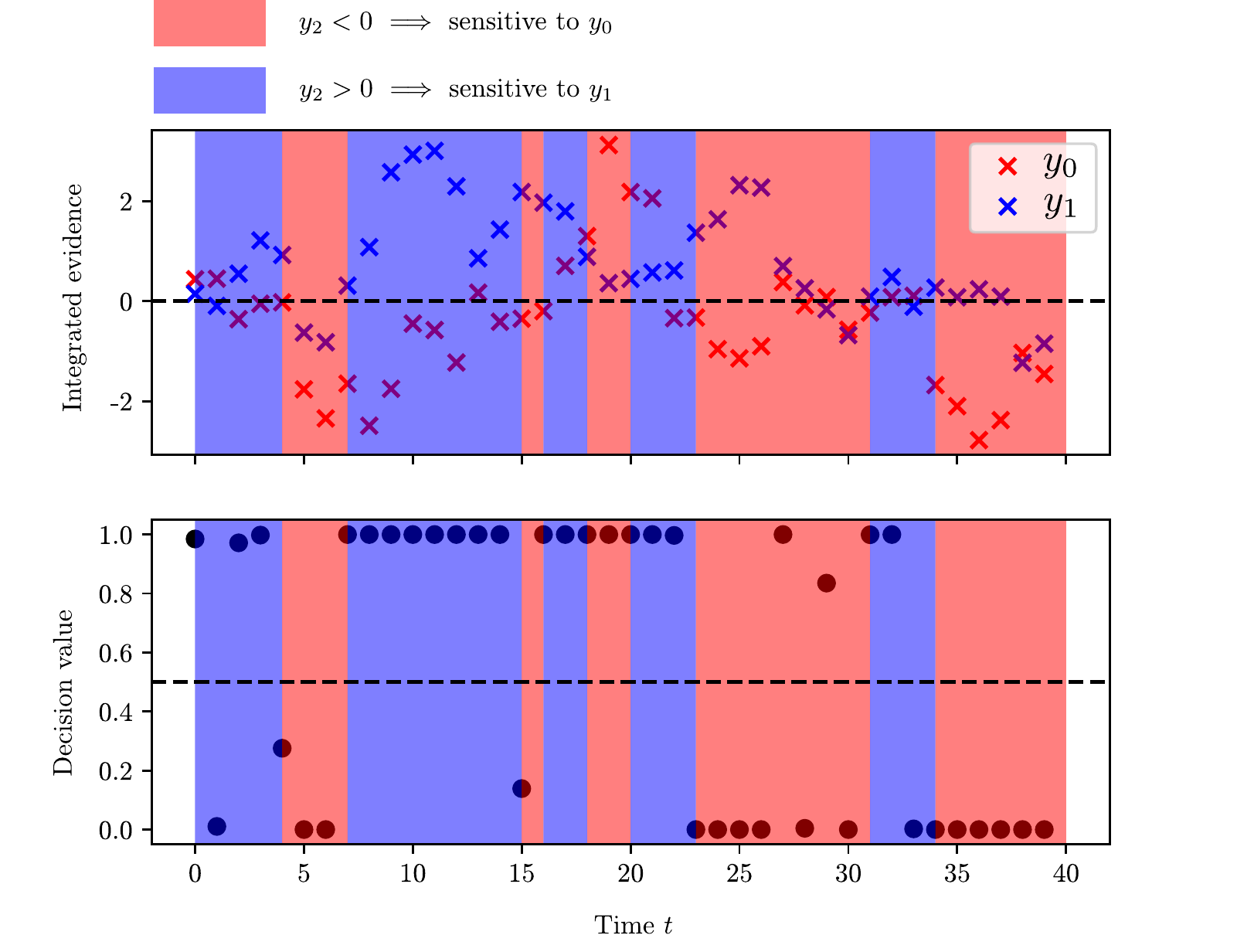}
    \caption{Output of an online context-dependent classifier. The task is the following: the network receives $D=3$ input channels; when the integral $y_2$ is negative, the network must output $0$ if $y_0<0$ and $1$ if $y_0>0$; when the integral $y_2$ is positive, the network must output $0$ if $y_1<0$, and $1$ if $y_1>0$. This result is obtained by training a sigmoidal decoding layer on the internal states of a fixed sigmoidal network pretrained through batch--SGD. \label{fig:mante}}
\end{figure}

In order to illustrate the versatility of the current-linear representations described in the main text, we implement a simple example of transfer learning to context-dependent selectivity, inspired by \citep{mante_context-dependent_2013}. The idea is the following: a pretrained $3$--channels integrator is used to integrate 3 time-series $x_0, \, x_1, \, x_2$ (respectively, the motion evidence, color evidence and contextual cue in the experiment described by \citep{mante_context-dependent_2013}) into their decaying integrals $y_0, \, y_1, \, y_2$, potentially with different decay constants. The cue integral $y_2$ is used to determine to which integral, $y_0$ or $y_1$, the network must be sensitive: when $y_2$ is negative, the network must output $0$ if $y_0<0$ and $1$ if $y_0>0$; when the integral $y_2$ is positive, the network must output $0$ if $y_1<0$, and $1$ if $y_1>0$.

To train this network, we first train the $3$--channels integrator using any of the methods described in the main text. We then use it as a fixed input transformer, mapping a $3$--dimensional time-series to a $n$--dimensional one (the state $\vh_t$ at any time-step). For each time-step, the value of the expected output is determined using the aforementioned rules on $\vy$, and the classification output is obtained as $out_t = (1+\exp^{-50(\vu\trs\vh_t-0.1)})^{-1}$. The trainable parameter of this new decoding layer is the vector $\vu$. It is easy to learn the value of $\vu$ through batch--SGD using the supervised learning procedure described here, and the resulting networks behave as shown in Figure \ref{fig:mante}.

% Appendices end up here

\end{document}